%% file: example.tex
\documentclass{article}
\usepackage[table,rgb]{xcolor}
\usepackage[preprint]{corl_2026} 
\usepackage[utf8]{inputenc} 
\usepackage[T1]{fontenc}    
\usepackage{hyperref}       
\usepackage{url}            
\usepackage{booktabs}       
\usepackage{amsfonts}       
\usepackage{nicefrac}       
\usepackage{microtype}      
\usepackage{multirow}
\usepackage{graphicx}  
\usepackage{booktabs}        
\usepackage{multirow}        
\usepackage{array}           
\usepackage{amsmath,amssymb} 
\usepackage{longtable}
\usepackage{listings}
\usepackage{tabularx}
\usepackage{wrapfig}
 \usepackage{tcolorbox}
\tcbuselibrary{breakable, skins, listings}

\definecolor{promptcomment}{rgb}{0.5,0.5,0.5}
\lstdefinestyle{pycodesm}{
  language         = Python,
  basicstyle       = \ttfamily\fontsize{7.5}{9.5}\selectfont,
  breaklines       = true,
  breakatwhitespace= false,
  keepspaces       = true,
  columns          = fullflexible,
  showstringspaces = false,
  morecomment      = [l][\color{promptcomment}\itshape]{\#},
}

\newcommand{\imbench}{\textsc{IMBench}\xspace}
\usepackage{xspace}

\newcommand{\task}[1]{\texttt{\small #1}}

\title{IMBench: A Benchmark for Intuitive Robotic Manipulation}

\author{
  Anurag Maurya \and
  Sukhvansh Jain \and
  Prajwal Avhad \and
  Gautham Balachandran \and
  Ziyi Zhou \and
  Atharva Kshirsagar \and
  Satyam Singh \and
  Bowen Li \and
  Rishabh Mukund \and
  Ritul Singh \and
  Jatin Vira \and
  Suvonil Chatterjee \and
  Devesh K. Jha\\[1em]
  \parbox{\textwidth}{\centering\textbf{Manav Robotics}, Bengaluru, India}
}

\begin{document}
\maketitle


\begin{abstract}
    Humans combine reasoning and motor control to solve complex manipulation tasks under diverse constraints. They build an understanding of the physical world that helps them convert reasoning into actions and quickly adapt to new scenes, tasks, and rules. We refer to
this capability as intuitive manipulation. Existing benchmarks fail to capture this integration: they evaluate physical reasoning in isolation from execution, or measure policy performance without requiring explicit reasoning. We introduce \imbench, a benchmark designed to evaluate intuitive manipulation as an
integrated capability spanning perception, physical reasoning, action generation, and iterative
execution. Our tasks require models to infer task-relevant physical structure and generate feasible action sequences under explicit constraints, including contact-rich manipulation, tool use, and multi-stage dependencies. We introduce a benchmark of $35$ tasks, $14$K filtered trajectories, and scalable tools for generating diverse scenarios. Experiments reveal a consistent gap: vision-language models show partial physical reasoning ability but fail to produce executable plans, while state-of-the-art vision-language-action models struggle to satisfy task constraints and generalize across scenarios. These results
identify intuitive manipulation as a missing axis in current foundation models and generalist
robot policies, and position \imbench as a step toward evaluating and enabling more integrated,
adaptive physical intelligence. \href{https://imbench.org}{\textbf{Project Website}}

\end{abstract}



\section{Introduction}

\begin{figure}[h]
    \centering
    \includegraphics[width=\textwidth]{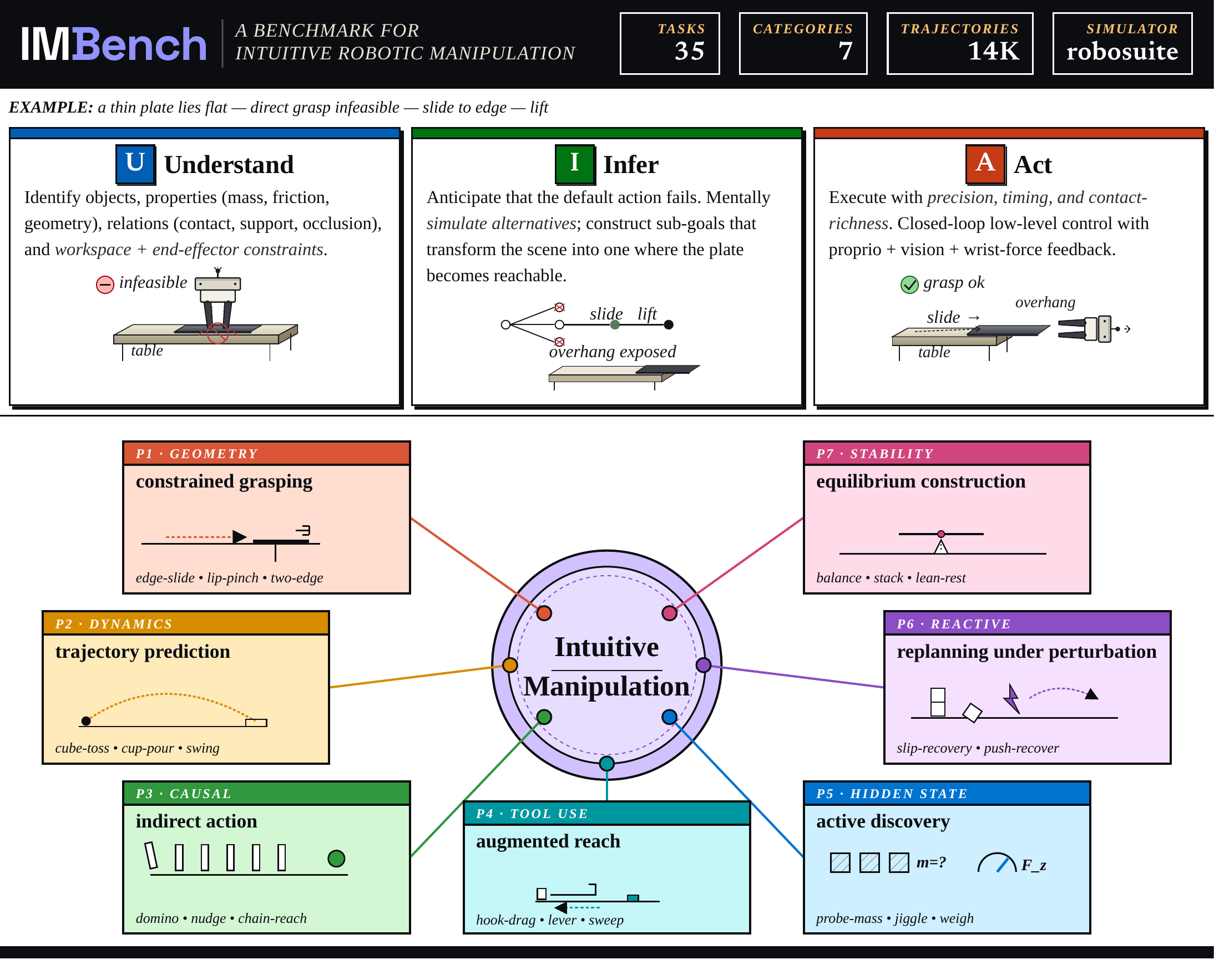}
    \caption{\textbf{IMBench — a benchmark for intuitive manipulation.}
    \textbf{Top:} Each task is decomposed into a three-stage cognitive loop. \emph{Understand} extracts latent physical properties from perception (e.g., a thin plate flush on a table affords no top-down grasp). \emph{Infer} simulates forward dynamics and selects sub-goals (e.g., slide-then-grasp). \emph{Act} executes a closed-loop interactive policy .
    }
    \label{fig:intro}
\end{figure}
 
Recent advances in robot learning have brought the field closer to generalist robots capable of performing a broad range of tasks across diverse environments~\citep{mccarthy2025towards}. However, a key gap remains: grounding behavior in the physical world the way humans do. Humans naturally reason about object properties, relationships, environmental constraints, and dynamics, using this understanding to guide flexible, adaptive actions~\citep{kubricht2017intuitive}. In contrast, current robotic systems struggle to capture and leverage such physical intuition, limiting their ability to generalize.

Consider a simple example (Fig.~\ref{fig:intro})(Top): a thin object lies flat on a table. A direct grasp is not feasible due to the geometry of the object. Humans, however, intuitively adapt, they slide the plate toward the edge of the table, enabling a stable grasp before lifting it. This behavior reflects more than passive understanding; it demonstrates the ability to convert physical understanding into goal-directed action. It involves reasoning about why a default action fails (\textit{Understand}), mentally simulating the effects of alternate actions (\textit{Infer}), and selecting a sequence that transforms an infeasible task into a feasible one (\textit{Act}). Two cognitive capacities support this behavior. The first is intuitive physics: the brain's approximate mental simulation of how objects, forces, and contacts behave~\citep{mccloskey1983intuitive}. The second is means-end reasoning, our ability to reason about goals, intentions, and constraints~\citep{griffiths2019doing}, and to act efficiently to achieve them~\citep{lake2017building}. We refer to their integration: the conversion of physical understanding into goal-directed action, as \emph{intuitive manipulation}.
 
Existing benchmarks probe the components of intuitive manipulation in isolation but rarely together. Physical-reasoning suites~\citep{bear2021physion, riochet2018intphys} such as PhysBench~\citep{chow2025physbench}, video-language QA benchmarks~\citep{johnson2017clevr, wu2024star, wang2024sok, goyal2017something, wang2025compositional} like CLEVRER~\citep{yi2019clevrer}, and embodied 3D-spatial QA benchmarks~\citep{du2024embspatial, ma2022sqa3d, jia2024sceneverse, lyu2024mmscan, yang2025thinking, wu2024vsp} like OpenEQA~\citep{majumdar2024openeqa} evaluate whether models can answer questions about object properties, relations, scene structure, and dynamics, but they stop at inference and do not synthesize executable action. Manipulation benchmarks: LIBERO~\citep{liu2023libero}, RoboCasa~\citep{nasiriany2024robocasa}, ManiSkill~\citep{maniskill_2021}, RLBench~\citep{james2019rlbench}, Colosseum~\citep{pumacay2024colosseum}, VLABench~\citep{zhang2025vlabench}, and OGBench~\citep{park2025ogbench} evaluate policy execution across diverse scenes and objects. Intuitive manipulation poses a complementary challenge: actions grounded in physical principles on tasks which heavily require them, an axis prior benchmarks and policies are not designed to isolate. Recent vision-language-action models (VLAs)~\citep{kim2024openvla, zitkovich2023rt2, team2024octo, zhen20243d} and related vision-language manipulation approaches~\citep{huang2023voxposer, niu2025llarva, wang2023mimicplay, zawalski2025robotic} generalize impressively across object identities and scene layouts, yet they underperform when the task structure itself requires this kind of physical reasoning, as we show empirically in \S\ref{sec:experiments}.
 
We introduce \textsc{IMBench}, a benchmark of 35 manipulation tasks built on robosuite~\citep{zhu2020robosuite} to evaluate intuitive manipulation across three capabilities: physical understanding, action proposal, and low-level execution. The tasks are intentionally non-trivial, requiring explicit physical reasoning to achieve task goals. They are organized around two key questions: (i) Can high-level planners infer task constraints from observations and propose feasible solutions? (ii) Can current physical-AI and foundation-model approaches translate physical understanding into low-level actions?

Our experiments reveal two key findings. First, vision-language models exhibit partial understanding of physical constraints. For instance, GPT-5.5 achieves $\approx74\%$ on constraint understanding, but drops to $\approx70\%$ on high-level planning, suggesting that the two capabilities are closely coupled. Second, performance degrades sharply when generating executable plans: even with correct high-level plans, action execution success falls to only $11\%$.  State-of-the-art policies, including diffusion policy~\citep{chi2023diffusionpolicy}, $\pi_{0.5}$~\citep{black2025pi05}, and GROOT~\citep{nvidia2024gr00t}, achieve low success rates on our physics-grounded tasks despite near-saturating performance on existing manipulation benchmarks such as LIBERO~\citep{liu2023libero}, highlighting the difficulty of \imbench.

\paragraph{Contributions:}This paper makes three primary contributions:
\begin{itemize}
    \item We present \textsc{IMBench}, a set of 35 robosuite manipulation tasks that test whether models understand physics and can use it to act, and we show where current foundation models and learned policies fail.
 
    \item We release a dataset and tools: $\approx$14K filtered trajectories, and scripts to generate more data supporting reproducible research for intuitive manipulation.
 
    \item We formalize intuitive manipulation through an \textit{Understand--Infer--Act} decomposition and evaluate it using a three-stage VLM protocol (constraint understanding $\rightarrow$ plan proposal $\rightarrow$ execution) together with closed-loop policy rollouts, enabling systematic measurement of the understanding-to-execution gap in current foundation models and policies.

\end{itemize}

\section{Related Work}

\paragraph{Physical Reasoning and Embodied Manipulation.}
Physical reasoning benchmarks study whether agents can infer object dynamics, contact interactions, and causal physical structure. PHYRE~\cite{bakhtin2019phyre}, I-PHYRE~\cite{li2024phyre} Kinetix~\cite{matthews2025kinetix}, and the Virtual Tools Game~\cite{allen2020rapid}, as well as vision-centric suites like Physion++~\cite{tung2023physion++}, ComPhy~\cite{chen2022comphy}, CRAFT~\cite{ates2022craft}, and CoPhy~\cite{baradel2019cophy}, alongside video-physics benchmarks~\cite{wu2016physics, patel2022cripp, girdhar2019cater, bansal2025videophy, duan2022pip} evaluate prediction and tool-use reasoning in simulated environments. However, these benchmarks terminate at inference, are usually MCQ-style and do not require physically executable robot behavior.
 
Manipulation benchmarks instead focus on policy learning and generalization across tasks and environments. LIBERO~\cite{liu2023libero}, CALVIN~\cite{mees2021calvin}, RoboCasa~\cite{nasiriany2024robocasa}, ManiSkill-HAB~\cite{shukla2025maniskill-hab}, BEHAVIOR-1K~\cite{li2024behavior1k}, FurnitureBench~\cite{heo2023furniturebench}, BiGym~\cite{chernyadev2025bigym}, and DittoGym~\cite{huang2024dittogym} study long-horizon manipulation, embodied interaction, and transfer across diverse scenes, often entangled with application-specific complexities as seen in household-assistance benchmarks like ALFRED~\citep{shridhar2020alfred} and AI2-THOR~\citep{kolve2017ai2}. VIMA~\cite{jiang2023vima} explores multimodal prompting for robotic manipulation, while large-scale datasets such as DROID~\cite{khazatsky2024droid} and BridgeData V2~\cite{walke2023bridgedatav2} emphasize scaling robot learning through diverse demonstrations. These benchmarks primarily evaluate trajectory reproduction, task completion, or policy generalization, but do not isolate whether success depends on reasoning about task-specific physical constraints. Most recently, KinDER~\cite{huang2026kinder} benchmarks kinematic and dynamic constraints and nonprehensile multi-object manipulation etc. IMBench is complementary with focus on reasoning in both physical and semantics axes with mixed grippers and bimanual setups, and adds hidden-state, reactive-replanning, and bimanual collaboration. KinDER also lacks force-information demonstrations. A comparison with existing benchmarks is provided in Appendix~\ref{app:related}.

\paragraph{Vision-Language-Action Models and Robot Planning.}
Recent Vision-Language-Action (VLA) models have significantly improved robot generalization across objects, scenes, and instructions. PaLM-E~\cite{driess2023palm}, RT-2~\cite{zitkovich2023rt2},
$\pi$-0.5~\cite{black2025pi05}, VLA-R1~\cite{ye2025vlar1}, and Gemini Robotics~\cite{geminirobotics2025} combine large-scale multimodal pretraining with action generation and reasoning capabilities. Parallel work on language-guided planning uses large language and vision-language models for embodied decision-making and replanning~\cite{Song_2023_ICCV, wang2024llm3, cherian2024llmphy, wang2023newton, yang2025embodiedbench}, including recent visual-prompting techniques like MOKA~\citep{fang2024moka} and PIVOT~\citep{nasiriany2024pivot}. Despite strong semantic generalization and recent physically-grounded VLM perception efforts~\cite{gao2024physically, chen2024spatialvlm, cheng2024spatialrgpt, rajani2020esprit}, these approaches still struggle when successful execution depends on reasoning about physical feasibility, or downstream action consequences. IMBench is designed to expose this limitation directly: tasks are constructed so that physically incorrect reasoning leads to execution failure even when high-level task intent is understood. Classical task and motion planning (TAMP) methods explicitly reason over symbolic and geometric constraints to ensure physically feasible manipulation~\cite{kaelbling2013integrated,garrett2021integrated,garrett2020pddlstream,zhao2024survey,lagriffoul2018platform}, but typically rely on hand-engineered abstractions and environment-specific formulations. \textbf{POSITIONING: } \imbench{} complements this line of work by evaluating whether modern learned policies and foundation models acquire such physical reasoning capabilities through embodied interaction.


\section{IMBench Task Suite}
\label{sec:tasks}

\textsc{IMBench} consists of 35 tasks built on \texttt{robosuite}~\citep{zhu2020robosuite}. Each task is designed around an \emph{inference bottleneck}: a hidden fact such as infeasibility, geometry, dynamics, causality, hidden state, or stability that the agent must infer from observations to achieve the goal reliably. Successful execution needs precise closed loop physical control. Together, these tasks test intuitive physical reasoning rather than simple imitation.

Tasks follow a Gymnasium-style interface with seeded resets. Episodes terminate on success (from simulator state), failure (e.g., object leaves workspace or forbidden contact), or a time limit, with sparse rewards ($1$ for success, $0$ otherwise). Observations include synchronized 2--4 RGB views, proprioception, gripper state, and wrist force/torque. Actions use a unified continuous space of 6-DoF end-effector deltas and gripper commands for both single-arm and bimanual setups. To ensure demonstration quality, episodes pass through a three-stage filtering pipeline: human annotators first score trajectories on a 1-5 scale (keeping only scores $\geq 4$), a VLM-based agent performs anomaly and consistency checks, and a final human review finalizes the decision. More details in Appendix~\ref{app:data}.

\begin{figure}[htbp]
    \centering
    \includegraphics[width=\textwidth]{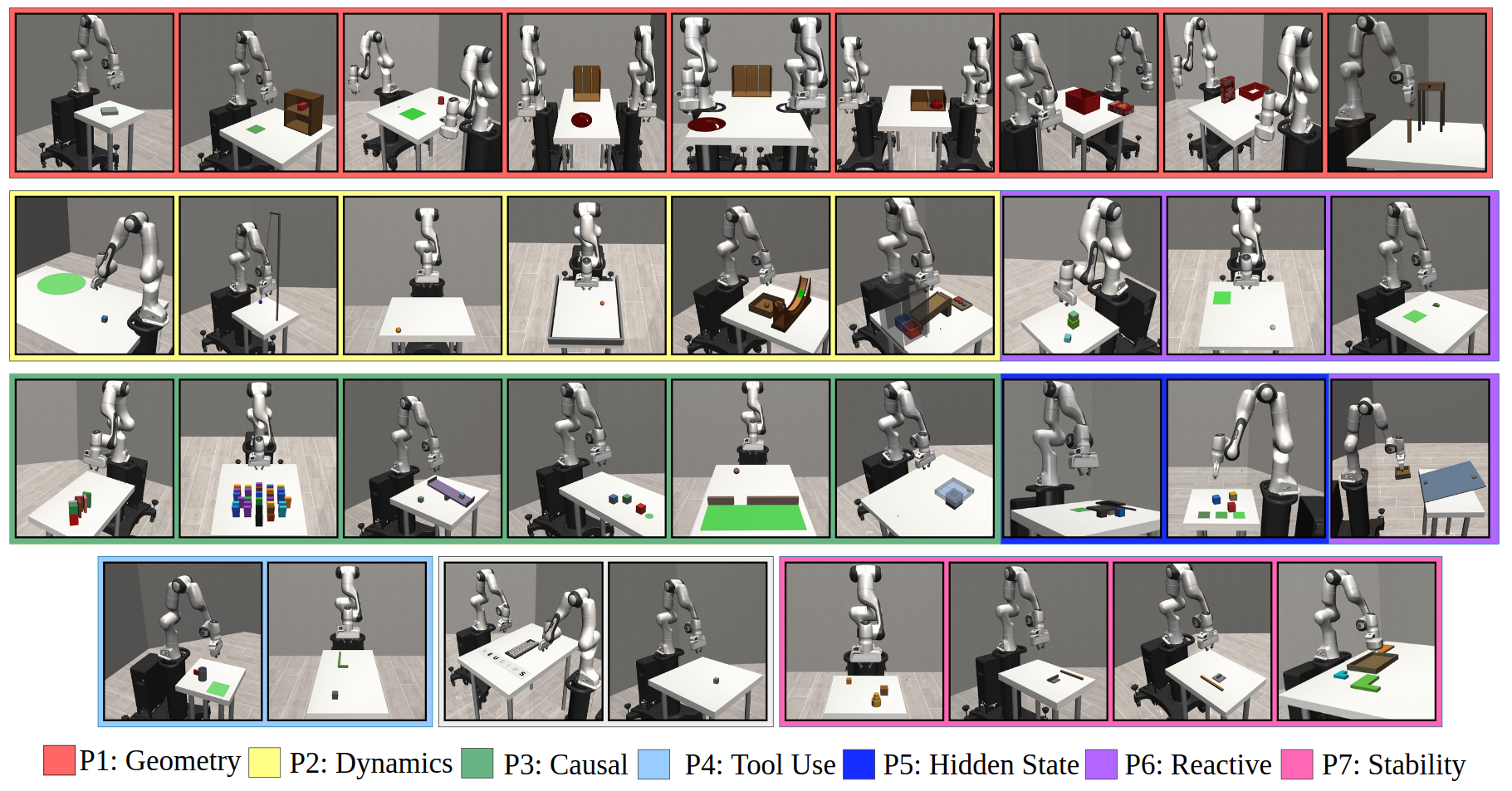}
    \caption{\textbf{IMBench: 35 tasks across seven categories.} Each targets a distinct inference bottleneck: geometry (P1), dynamics (P2), causal chains (P3), tool use (P4), hidden state (P5), reactive replanning (P6), stability (P7); two miscellaneous tasks in grey (bottom row).}
    \label{fig:tasks}
\end{figure}

\subsection{Categories and inference bottlenecks}
\label{sec:categories}
We group the 35 tasks into seven categories, each defined by the intuition getting invoked (Fig.~\ref{fig:tasks}); per-task details appear in Appendix~\ref{app:percat}.
\textbf{P1 -- Geometry and constrained grasping (9 tasks)} require the agent to make objects graspable or reason about geometry (e.g., edge-sliding thin plates, suction-to-jaw handovers, lateral rack entry), testing affordance reasoning and sub-goal ordering.
\textbf{P2 -- Dynamics and trajectory prediction (7 tasks)} places goals beyond static reach or moving objects (tossing, pendulum grasping, rolling-ball interception with and without bounces, color-routed sorting), probing forward simulation.
\textbf{P3 -- Causal and indirect action (6 tasks)} makes the target unreachable or contact-forbidden, so action must transit an intermediary chain (domino cascades with sabotaged distractors, seesaw torque balance under unknown mass, transitive pushing), probing causal-chain construction and structural reasoning.
\textbf{P4 -- Tool use and augmented reach (2 tasks)} requires using a tool or actions \emph{before} the final grasp, probing multi-stage ordering under embodiment limits.
\textbf{P5 -- Hidden state and active discovery (2 tasks)} hides the decision-relevant variable from passive vision; identity only by displacing occluders---probing information-gathering actions and multi-modal grounding.
\textbf{P6 -- Reactive replanning (3 tasks)} injects mid-execution disturbances (gripper slip, knocked stack, teleporting goal), probing online failure detection and replanning.
\textbf{P7 -- Stability and equilibrium (4 tasks)} hinges feasibility on static balance (mixed-shape stacking, rod balancing on millimeter ridges, zero-gap L-piece packing), probing support-polygon reasoning and precision under low tolerance.
\textbf{Miscellaneous (2 tasks)} adds \textit{mirror-pick-place}, which rotates camera frames by $180^\circ$ to decouple observation and action frames, and \textit{keyboard-typing}, a bimanual task where one arm partially finishes the goal, probing partial-completion tracking and coordination. The full task list and additional details are provided in Appendix~\ref{app:tasks}.
 \\


\input{experiments}

\section{Limitations}
\label{sec:limitations}
 IMBench is currently limited to simulation and evaluates only Franka (widely-supported embodiment) with parallel-jaw and suction grippers, excluding dexterous hands, mobile platforms, and humanoids. Task descriptions are fixed, tasks are low-to-mid horizon, and deformable object manipulation is not considered. While we release force-torque and tactile signals, the evaluated baselines do not use tactile or force-feedback modalities; principled integration of such signals remains an open direction that may improve performance. Finally, we focus on end-to-end vision-language-action baselines without privileged state information, emphasizing generalist agents and policies rather than task-specific engineered systems such as TAMP or long-horizon planners.


\section{Conclusion}
\label{sec:conclusion}
We introduce \imbench, a benchmark of 35 manipulation tasks that test whether models can turn physical understanding into action. Tasks are organized around an Understand–Infer–Act loop and grouped by the intuition each one probes. We release \textasciitilde14K curated demonstrations, a three-stage curation pipeline, and a scripted-demonstration generator, and we evaluate both high-level reasoning and low-level execution on current foundation models.

Three findings stand out. Frontier VLMs identify task constraints with moderate accuracy but propose valid plans less often. State-of-the-art VLAs score near zero zero-shot and perform poorly overall even after in-domain fine-tuning. And under OOD shifts along each task's target physical axis, all policies drop sharply. We position intuitive manipulation as a missing axis of capability, orthogonal to the dimensions current benchmarks reward, and release \textsc{IMBench} as a measurement instrument to help the field localize where intuitive physical intelligence lives in current systems.


\clearpage
\acknowledgments{If a paper is accepted, the final camera-ready version will (and probably should) include acknowledgments. All acknowledgments go at the end of the paper, including thanks to reviewers who gave useful comments, to colleagues who contributed to the ideas, and to funding agencies and corporate sponsors that provided financial support.}


\bibliography{example}  
\newpage
\appendix
\input{appendix}

\end{document}

%% file: experiments.tex

\section{Experiments}
\label{sec:experiments}

The experimental section is organized around three questions:

\begin{itemize}
\item[\textbf{Q1.}] \textit{Constraint understanding.}
  Given a scene observation(s) and a required goal, can frontier
  vision-language models (VLMs) correctly identify the physical
  constraints: infeasible grasps, hidden state, causal
  dependencies, that prevents naive actions from succeeding?

\item[\textbf{Q2.}] \textit{High-level action proposal.}
  Conditioned on a (possibly correct) understanding of the scene, can the
  same models translate that understanding into high-level plan, expressed as a verifiable and executable sequence of sub-goals?
  
\item[\textbf{Q3.}] \textit{Execution and generalization.}
Can current visuomotor policies execute physics grounded plans on \imbench and generalize beyond the training distribution?
\end{itemize}

Q1--Q2 examine \emph{high-level} reasoning, while Q3 focuses on
\emph{low-level} execution and tests whether policies generalize along
the physical factors each task targets.

\subsection{High-level reasoning with VLMs (Q1, Q2)}
\label{sec:vlm}
We evaluate five high-level reasoners:
(i)~\textbf{GPT-5.5}
(ii)~\textbf{GPT-5.4-mini},
(iii)~\textbf{Gemma-4},
(iv)~\textbf{Claude~Haiku-4.5},
and (v)~\textbf{Claude-Sonnet-4.6}. All models use a shared chain-of-thought prompting interface to elicit structured traces for human evaluation.  For each seed, we render multi-view observations and provide the task description and controller interface (DoF, gripper modes, observables). A small set of temporally spaced frames is given per task. From each response we extract two outputs: \textbf{1) Stage~1: Constraint understanding}
Structured responses to probes (objects, contacts, reachability, infeasibility, hidden state, causal preconditions), evaluated via a task-specific rubric (given to the human evaluator). Success requires identifying all task-critical constraints and interactions. \textbf{2) Stage~2: High-level plan}
A sequence of sub-goals $(g_1,\ldots,g_T)$ (e.g., \textsc{Grasp}, \textsc{Slide}, \textsc{Toss}), composed by the model without hints. This tests its ability to form task-aligned plans consistent with key constraints. Restricting outputs to primitives reduces the model to MCQ style selection instead of generating new behaviors. Results are aggregated over five temperature settings. Details are provided in Appendix~\ref{sec:vlm_evaluation}.

\begin{table}[t]
\centering
\small
\caption{\textbf{VLM evaluation across three stages.} \textbf{Stage~1} reports constraint-understanding success (\%), \textbf{Stage~2} reports high-level plan correctness (\%), and \textbf{Stage~3} reports closed-loop execution success (\%). A Stage~2 plan is marked correct if a human verifies that the proposed sub-goals satisfy task constraints and can achieve task success. Best per column within each stage is shown in \textbf{bold}.}
\label{tab:vlm-stages}
\setlength{\tabcolsep}{5pt}
\resizebox{\textwidth}{!}{%
\begin{tabular}{lcccccccccc}
\toprule
Method & P1 & P2 & P3 & P4 & P5 & P6 & P7 & M & Mean \\
\midrule
\multicolumn{10}{l}{\textit{Stage 1: Constraint understanding}} \\
\midrule
Gemma 4           & $51.0\!\pm\!30.6$ & $65.2\!\pm\!29.6$ & $79.3\!\pm\!18.5$ & $86.0\!\pm\!0.0$ & $90.0\!\pm\!0.0$ & $50.6\!\pm\!31.1$ & $37.5\!\pm\!10.1$ & $0.0\!\pm\!0.0$ & $57.8\!\pm\!32.0$ \\
Claude-Haiku 4.5  & $63.1\!\pm\!33.5$ & $85.4\!\pm\!10.8$ & $80.7\!\pm\!22.1$ & $70.0\!\pm\!0.0$ & $83.8\!\pm\!16.2$ & $36.7\!\pm\!4.7$ & $35.0\!\pm\!30.2$ & $2.0\!\pm\!2.0$ & $63.6\!\pm\!32.8$ \\
Claude Sonnet 4.6 & $65.5\!\pm\!33.0$ & $\mathbf{93.1}\!\pm\!5.0$ & $90.0\!\pm\!10.0$ & $90.0\!\pm\!0.0$ & $\mathbf{97.5}\!\pm\!2.5$ & $62.8\!\pm\!27.8$ & $65.0\!\pm\!27.3$ & $0.0\!\pm\!0.0$ & $\mathbf{74.5}\!\pm\!31.2$ \\
GPT-5.4-Mini      & $55.2\!\pm\!31.0$ & $76.7\!\pm\!24.3$ & $82.2\!\pm\!14.3$ & $90.0\!\pm\!0.0$ & $95.0\!\pm\!5.0$ & $62.2\!\pm\!32.5$ & $46.7\!\pm\!23.9$ & $5.0\!\pm\!5.0$ & $64.8\!\pm\!31.6$ \\
GPT-5.5           & $\mathbf{71.3}\!\pm\!35.4$ & $75.8\!\pm\!30.1$ & $\mathbf{93.7}\!\pm\!9.0$ & $\mathbf{90.6}\!\pm\!4.6$ & $94.2\!\pm\!4.1$ & $\mathbf{56.7}\!\pm\!40.3$ & $\mathbf{80.5}\!\pm\!15.5$ & $\mathbf{8.3}\!\pm\!8.3$ & $74.1\!\pm\!33.0$ \\
\midrule
\multicolumn{10}{l}{\textit{Stage 2: Plan correctness}} \\
\midrule
Gemma 4           & $46.7\!\pm\!31.3$ & $53.9\!\pm\!37.1$ & $66.5\!\pm\!21.7$ & $52.0\!\pm\!0.0$ & $90.0\!\pm\!0.0$ & $17.8\!\pm\!5.7$ & $11.7\!\pm\!12.8$ & $2.0\!\pm\!2.0$ & $45.3\!\pm\!34.3$ \\
Claude-Haiku 4.5  & $56.7\!\pm\!36.4$ & $72.6\!\pm\!27.8$ & $67.2\!\pm\!26.3$ & $24.0\!\pm\!0.0$ & $83.8\!\pm\!16.2$ & $48.9\!\pm\!9.6$ & $28.3\!\pm\!25.2$ & $2.0\!\pm\!2.0$ & $55.7\!\pm\!33.9$ \\
Claude Sonnet 4.6 & $55.2\!\pm\!33.9$ & $\mathbf{89.4}\!\pm\!9.2$ & $84.2\!\pm\!14.8$ & $25.0\!\pm\!0.0$ & $\mathbf{97.5}\!\pm\!2.5$ & $50.6\!\pm\!28.3$ & $39.2\!\pm\!31.0$ & $0.0\!\pm\!0.0$ & $64.1\!\pm\!34.6$ \\
GPT-5.4-Mini      & $45.2\!\pm\!29.9$ & $58.0\!\pm\!36.1$ & $76.9\!\pm\!19.1$ & $74.0\!\pm\!0.0$ & $91.5\!\pm\!3.5$ & $30.6\!\pm\!19.7$ & $14.6\!\pm\!8.1$ & $5.0\!\pm\!5.0$ & $50.0\!\pm\!34.4$ \\
GPT-5.5           & $\mathbf{69.0}\!\pm\!35.4$ & $61.3\!\pm\!40.8$ & $\mathbf{93.0}\!\pm\!9.9$ & $\mathbf{75.9}\!\pm\!5.9$ & $94.2\!\pm\!4.1$ & $\mathbf{56.7}\!\pm\!40.3$ & $\mathbf{80.0}\!\pm\!15.2$ & $\mathbf{8.3}\!\pm\!8.3$ & $\mathbf{69.5}\!\pm\!35.7$ \\

\midrule
\multicolumn{10}{@{}l}{\textit{Stage 3: Action, uses GPT 5.5}} \\
\midrule
IMBAgent     & 0.0  & 0.0 & 75.0  & 0.0 & 0.0 & 0.0 & 0.0 & 15.0 & 11.3 \\
IMBAgent-obj & 0.0  & 0.0 & \textbf{100.0} & 0.0 & 0.0 & 0.0 & 0.0 & \textbf{50.0} & \textbf{18.8} \\
\bottomrule
\end{tabular}
}
\end{table}

\subsubsection{Results: understanding (Stage~1)}

Table~\ref{tab:vlm-stage1} shows that frontier VLMs
describe most constraints competently, but the failures concentrate on a
small, consistent set of \emph{tasks} rather than on whole categories
(Table~\ref{tab:vlm-stage1}). Several tasks are understood almost perfectly by
all models, including \task{domino-single}, \task{indirect-push},
\task{linear-intercept}, \task{bounce-intercept},
\task{pendulum-grasp}, and \task{sheltered-grasp}
($90$--$100\%$). These strong results raise category-level scores for
geometry (P1), dynamics (P2), and causal action (P3), with
Sonnet~4.6 ($74.5\%$) and GPT-5.5 ($74.1\%$) achieving the best overall
Stage-1 performance (Table~\ref{tab:vlm-stages}).

The weakest performance appears on tasks such as
\task{cracker-box-jaw}, \task{cracker-box-suction},
\task{shape-stack}, and \task{slip-recovery}, where models recognize the
objects but fail to identify the key grasp or stability constraint.
Interestingly, \task{balance-hard} often performs better than
\task{balance-medium} because its prompt explicitly encourages
center-of-mass reasoning. Stability and equilibrium tasks (P7) remain the weakest overall
category ($37$--$80\%$) escept miscellaneous category. Additional analysis is provided in
Appendix~\ref{app:insights_stage1}.

\subsubsection{Results: action proposal (Stage~2)}

The understanding-to-plan gap is itself task-specific. The largest drop from understanding to planning appears in tasks that
require correctly ordered actions, while static geometry tasks largely retain their Stage-1 performance (Table~\ref{tab:vlm-stage2}). \task{slide-catch} is the clearest example: despite Stage-1 scores of
$88$--$95\%$, most models fail completely in Stage~2 because their plans miss important timing and tracking steps. Similar failures appear in \task{tool-retrieve}, \task{cup-inversion},
\task{stack-collapse-recovery}, and \task{cup-extract}. In contrast, tasks such as \task{gap-funnel},
\task{sheltered-grasp}, \task{ramp-sort}, and \task{domino-single}
transfer almost perfectly from understanding to planning
($85$--$100\%$). GPT-5.5 maintains strong performance on geometry, dynamics, and causal-action categories, while stability reasoning (P7)
remains inconsistent across models. Overall, the main challenge is action sequencing rather than recognizing the constraint itself
(Appendix~\ref{subsec:stage2_results}).

\subsection{Closed-Loop Manipulation Execution (Q3)}
\label{sec:policies}

\subsubsection{Reasoning-to-Action Execution (Stage~3)}

We extend the above agentic framework to translate reasoning into executable manipulation actions. In Stage~3, we use a closed-loop \textbf{ReAct (Reasoning + Acting)} framework where the VLM receives multi-view images, proprioception, and force-torque signals, then selects predefined action primitives with iterative feedback after each step. We evaluate GPT-5.5 in both vision-only and privileged object-centric settings across representative tasks from all categories. Due to the high cost of interactive evaluation (A complete Stage~3 evaluation across 16 tasks costs  approximately \$1800), we benchmark only GPT-5.5 as an agent, reported as \texttt{IMBAgent} and \texttt{IMBAgent-obj} in Table~\ref{tab:vlm-stages}. To provide broader insights, we additionally evaluate end-to-end visuomotor policies on \imbench. Further implementation details are provided in Appendix~\ref{app:stage3}.

Closed-loop execution exposes the largest reasoning-to-action gap.
Across sixteen evaluated tasks, GPT-5.5 achieves non-zero success on only
three (Table~\ref{tab:vlm-stage3-gpt55}). \task{domino-single} succeeds in
all runs because corrective pushes tolerate imprecise contact. Privileged
object poses improve \task{domino-select} and
\task{mirror-pick-place}, increasing the mean success rate from
$11.3\%$ to $18.8\%$. However, all tasks requiring precise alignment,
timing, tool use, hidden-state reasoning, or balancing score $0\%$ in both
settings. In some cases, privileged poses even hurt reasoning: on
\task{mirror-pick-place}, the agent follows reported coordinates directly
without applying the required mirror transformation. Per-task rollout
analysis is provided in Appendix~\ref{app:insights_stage3}.

\subsubsection{End-to-end manipulation policies}
We evaluate three end-to-end visuomotor policies, trained on the 200 human demonstration trajectories.  
\textbf{Diffusion Policy (DP)} is trained from scratch for each task, while \textbf{$\pi_{0.5}$} and \textbf{GR00T~N\,1.5} (vision--language--action policies) are evaluated in zero-shot and finetuned settings. We evaluate two regimes: (1)~\textbf{Zero-shot (ZS)}, using publicly available checkpoints without task-specific finetuning applied only to $\pi_{0.5}$ and GR00T~N\,1.5; and (2)~\textbf{expert-demonstration training}, where all three approaches are trained/finetuned on curated \imbench data. For tasks where teleoperation is infeasible, data is collected using manually designed scripted policies. Training configurations are kept as close as possible to the original implementations. Full architecture and training details are in Appendix~\ref{app:training}.

Zero-shot VLA performance is extremely low ($\le\!0.02$ mean), and useful
behavior only emerges after task-specific training
(Table~\ref{tab:per-policy-results}). Diffusion Policy performs best on
mirror reasoning and contact-tolerant tasks, including
\task{mirror-pick-place}, \task{domino-select},
\task{occluder-push}, and \task{slip-recovery}. Finetuned $\pi_{0.5}$
mainly succeeds on balancing tasks such as
\task{balance-medium} and \task{balance-hard}. Several tasks remain unsolved by all policies, including
\task{cup-inversion}, \task{shape-stack},
\task{stack-collapse-recovery}, \task{packing},
\task{mass-sort}, and both cracker-box grasp variants.
Detailed task-level analysis is provided in
Appendix~\ref{app:insights_vismotor}.

\begin{table}[htbp]
\centering
\small
\caption{\textbf{Policy success rates (\%)} on \imbench, averaged over
tasks within each category. Rollouts are over 20 seeds and 5 rollouts each. Bold marks the best entry per row. ZS = zero-shot pretrained
checkpoint; FT = finetuned on teleoperation data. Diffusion Policy (DP)
is trained from scratch (Full-training).}
\label{tab:policy-main}
\begin{tabularx}{\textwidth}{@{}lXcccccc@{}}
\toprule
& & \multicolumn{2}{c}{\textbf{$\pi_{0.5}$}}
  & \multicolumn{2}{c}{\textbf{GR00T~1.5}}
  & \multicolumn{1}{c}{\textbf{DP}} \\
\cmidrule(lr){3-4}\cmidrule(lr){5-6}
& & \textbf{ZS} & \textbf{FT} & \textbf{ZS} & \textbf{FT} & \textbf{Full-training} \\
\midrule
  P1 & Geometry and constrained grasp   & 0.00 & 0.05 & 0.00 & 0.00 & \textbf{0.10} \\
  P2 & Dynamics and trajectory          & 0.01 & 0.09 & 0.00 & 0.02 & \textbf{0.12} \\
  P3 & Causal and indirect action       & 0.08 & 0.29 & 0.03 & 0.11 & \textbf{0.37} \\
  P4 & Tool use and augmented reach     & 0.00 & 0.00 & 0.00 & 0.00 & 0.00 \\
  P5 & Hidden state and discovery       & 0.00 & 0.08 & 0.00 & 0.00 & \textbf{0.25} \\
  P6 & Reactive replanning              & 0.00 & 0.17 & 0.00 & 0.00 & \textbf{0.24} \\
  P7 & Stability and equilibrium        & 0.00 & \textbf{0.32} & 0.00 & 0.01 & 0.12 \\
  \midrule
  & Miscellaneous                       & 0.01 & 0.06 & 0.00 & 0.02 & \textbf{0.65} \\
  \midrule
  & \textbf{Mean}                       & 0.01 & 0.15 & 0.00 & 0.02 & \textbf{0.24} \\
\bottomrule
\end{tabularx}
\end{table}

\subsection{Out-of-distribution generalization}
\label{sec:ood}

Changing the main physical factor of each task significantly reduces
performance (Table~\ref{tab:ood}) on tasks with the highest
in-distribution scores. This suggests that policies rely heavily on
memorized patterns rather than robust physical understanding.
For example, \task{balance-medium} drops from $0.71$ to $0.12$ after a
center-of-mass perturbation, while \task{keyboard-typing} falls from
$0.32$ to $0.04$. Contact-tolerant tasks such as \task{domino-single} remain relatively
robust. A few tasks show small improvements under perturbation, but overall performance degrades.

\begin{table}[t]
\centering
\small
\caption{\textbf{OOD generalization (task-wise).} Success rate under in-distribution (ID) and out-of-distribution (OOD) settings, with absolute drop $\Delta$.}
\label{tab:ood}
\begin{tabular*}{\textwidth}{@{\extracolsep{\fill}}llccccccccc@{}}
\toprule
& & \multicolumn{3}{c}{\textbf{$\pi_{0.5}$}}
  & \multicolumn{3}{c}{\textbf{GR00T~1.5}}
  & \multicolumn{3}{c}{\textbf{DP}} \\
\cmidrule(lr){3-5}\cmidrule(lr){6-8}\cmidrule(lr){9-11}
& & \textbf{ID} & \textbf{OOD} & $\boldsymbol{\Delta}$
  & \textbf{ID} & \textbf{OOD} & $\boldsymbol{\Delta}$
  & \textbf{ID} & \textbf{OOD} & $\boldsymbol{\Delta}$ \\
\midrule
P1  & sheltered-grasp       & 0.10 & 0.00 & -0.10 & 0.00 & 0.00 & 0.00 & 0.30 & 0.00 & -0.30 \\
P2  & pendulum-grasp        & 0.28 & 0.40 & +0.12 & 0.08 & 0.04 & -0.04 & 0.35 & 0.00 & -0.35 \\
P3  & domino-single         & 0.73 & 0.52 & -0.21 & 0.25 & 0.15 & -0.10 & 0.70 & 0.60 & -0.10 \\
P4  & tool-retreive         &   0.00   &  0.00   &   0.00    &   0.00   &  0.00    &  0.00     &   0.00   &   0.00   &   0.00    \\
P5  & occluder-push         & 0.15 & 0.08 & -0.07 & 0.00 & 0.00 & 0.00 & 0.50 & 0.40 & -0.10 \\
P6  & slip-recovery         & 0.22 & 0.08 & -0.14 & 0.00 & 0.00 & 0.00 & 0.44 & 0.56 & +0.12 \\
P7  & balance-medium        & 0.71 & 0.12 & -0.59 & 0.02 & 0.00 & -0.02 & 0.28 & 0.04 & -0.24 \\
M   & keyboard-typing       & 0.13 & 0.00 & -0.13 & 0.04 & 0.00 & -0.04 & 0.32 & 0.04 & -0.28 \\

\bottomrule
\end{tabular*}
\end{table}

\subsection{Discussion}
\label{sec:discussion}

Our results highlight three main findings. First, VLMs are much better at
recognizing physical constraints than converting them into executable
action plans. Tasks with simple geometric reasoning transfer well from
understanding to planning, while tasks requiring ordered or timed actions
often fail. Second, policy learning succeeds mainly on contact-tolerant tasks and a
small number of structured behaviors such as mirror reasoning. Tasks
requiring precise alignment, grasp mechanics, reorientation, or balancing
remain unsolved across nearly all policies. Third, learned behaviors generalize poorly outside the training
distribution. Even the strongest balancing policy drops sharply under
small center-of-mass changes, and category-level performance remains low
across geometry, dynamics, and stability tasks. Together, these results
suggest that intuitive manipulation remains a missing capability axis that
is not captured by current physical reasoning or scene understanding
benchmarks.

%% file: appendix.tex

\appendix

\section{Related Benchmarks}
\label{app:related}

\imbench differs from prior benchmarks by explicitly evaluating the full pipeline from physical reasoning to executable manipulation. It separately measures VLM constraint understanding, plan generation, and closed-loop execution while covering all five key physical reasoning categories. In addition, it introduces bimanual coordination tasks with expert demonstrations, enabling evaluation of both reasoning and control. We additionally release tactile and force/torque data to support multimodal manipulation research and improve performance on \imbench.

\begin{table*}[h!]
\centering
\small
\setlength{\tabcolsep}{3pt}
\renewcommand{\arraystretch}{1.1}
\caption{\textbf{Related Benchmarks.} \imbench is the only benchmark that (i) focuses on physical reasoning, (ii) evaluates VLM constraint understanding and plan generation \emph{separately} from closed-loop policy execution, (iii) covers all five physical reasoning categories (geometry, dynamics, causal chains, hidden state, reactive replanning), and (iv) provides bimanual coordination tasks with demonstrations.}
\label{tab:related}
\newcommand{\cmark}{\textcolor{green!60!black}{\checkmark}}
\newcommand{\xmark}{\textcolor{red}{$\times$}}
\resizebox{\linewidth}{!}{%
\begin{tabular}{@{}lccccccccc@{}}
\toprule
\textbf{Benchmark}
  & \shortstack[c]{\textbf{Phys.\ Reasoning}\\\textbf{Focus}}
  & \shortstack[c]{\textbf{Sep.\ VLM +}\\\textbf{Policy Eval}}
  & \shortstack[c]{\textbf{Geometry /}\\\textbf{Affordance}}
  & \shortstack[c]{\textbf{Dynamics /}\\\textbf{Prediction}}
  & \shortstack[c]{\textbf{Causal /}\\\textbf{Indirect}}
  & \shortstack[c]{\textbf{Hidden}\\\textbf{State}}
  & \shortstack[c]{\textbf{Reactive}\\\textbf{Replanning}}
  & \shortstack[c]{\textbf{Bimanual}\\\textbf{Coordination}}
  & \textbf{Demonstrations} \\
\midrule
PHYRE~\cite{bakhtin2019phyre}              & \cmark & \xmark & \xmark & \cmark & \cmark & \xmark & \xmark & \xmark & \xmark \\
PhysBench~\cite{chow2025physbench}         & \cmark & \xmark & \xmark & \cmark & \xmark & \xmark & \xmark & \xmark & \xmark \\
CLEVRER~\cite{yi2019clevrer}               & \cmark & \xmark & \xmark & \cmark & \cmark & \xmark & \xmark & \xmark & \xmark \\
LIBERO~\cite{liu2023libero}                & \xmark & \xmark & \xmark & \xmark & \xmark & \xmark & \xmark & \xmark & \cmark \\
RLBench~\cite{james2019rlbench}            & \xmark & \xmark & \cmark & \xmark & \xmark & \xmark & \xmark & \xmark & \cmark \\
RoboCasa~\cite{nasiriany2024robocasa}      & \xmark & \xmark & \cmark & \xmark & \xmark & \xmark & \xmark & \xmark & \cmark \\
ManiSkill-HAB~\cite{shukla2025maniskill-hab}& \xmark & \xmark & \xmark & \xmark & \xmark & \xmark & \xmark & \xmark & \cmark \\
VLABench~\cite{zhang2025vlabench}          & \xmark & \xmark & \xmark & \xmark & \cmark & \xmark & \xmark & \xmark & \cmark \\
Colosseum~\cite{pumacay2024colosseum}      & \xmark & \xmark & \cmark & \xmark & \xmark & \xmark & \xmark & \xmark & \cmark \\
BiGym~\cite{chernyadev2025bigym}            & \xmark & \xmark & \xmark & \xmark & \xmark & \xmark & \xmark & \cmark & \cmark \\
VLMgineer~\cite{vlmgineer2025}             & \cmark & \cmark & \cmark & \xmark & \xmark & \xmark & \xmark & \xmark & \xmark \\
KinDER~\cite{huang2026kinder}              & \cmark & \xmark & \cmark & \cmark & \cmark & \xmark & \xmark & \xmark & \cmark \\

\midrule
\textbf{\imbench~(Ours)}                   & \cmark & \cmark & \cmark & \cmark & \cmark & \cmark & \cmark & \cmark & \cmark \\
\bottomrule
\end{tabular}}
\end{table*}

\section{Task Suite --- Full Specifications}
\label{app:tasks}
This appendix gives the full specification of every \textsc{IMBench} task. Each entry lists: phenomenon category, embodiment, end-effector configuration, the physical axis the task isolates, the success predicate, the reset-time randomization ranges, the canonical sub-goal sequence used by the Stage-2 oracle, the demonstration counts after curation, and the mean episode length. We release a subset of the dataset (10 episodes) via Hugging Face for review purposes. To facilitate easy inspection, we provide utilities that allow reviewers to directly visualize the dataset through a web link without requiring any installation. The complete benchmark and full dataset will be released upon acceptance. Each task description below includes an associated link that redirects to the LeRobot visualizer for interactive inspection.

\subsection{Master catalog}
\label{app:catalog}

\begin{table}[h]
\centering
\small
\caption{Master catalog of \textsc{IMBench} tasks. ``Demos'' lists curated counts after the three-stage filtering pipeline. A dash (--) in the teleop column denotes a task for which teleoperation is infeasible with the SpaceMouse interface.}
\label{tab:master-catalog}
\begin{tabular}{llllrr}
\toprule
ID & Task & Cat. & Embodiment / EE & Demos (teleop / scripted) & Mean ep.\ (s) \\
\midrule
T01 & edge-slide              & P1   & 1-arm / jaw        & 200 / 200 & 12.4 \\
T02 & recover-peg-insert      & P1   & 1-arm / jaw        & 200 / 200 & 14.1 \\
T03 & sheltered-grasp         & P1   & 1-arm / jaw        & 200 / 200 & 9.8  \\
T04 & cup-inversion           & P1   & het.\ bi / suc+jaw & 200 / 200 & 18.6 \\
T05 & plate-shelf-hobi        & P1   & hom.\ bi / jaw+jaw & 200 / 200 & 21.3 \\
T06 & plate-shelf-hetbi       & P1   & het.\ bi / suc+jaw & 200 / 200 & 19.7 \\
T07 & bowl-table              & P1   & het.\ bi / suc+jaw & 200 / 200 & 17.2 \\
T08 & cracker-box-suction     & P1   & 1-arm / suction    & 200 / 200 & 11.7 \\
T09 & cracker-box-jaw         & P1   & 1-arm / jaw        & 200 / 200 & 13.2 \\
T10 & cube-toss               & P2   & 1-arm / jaw        & --  / 400 & 4.6  \\
T11 & pendulum-grasp          & P2   & 1-arm / jaw        & --  / 400 & 7.3  \\
T12 & linear-intercept        & P2   & 1-arm / jaw        & --  / 400 & 5.1  \\
T13 & bounce-intercept        & P2   & 1-arm / jaw        & --  / 400 & 6.8  \\
T14 & galileo-ramp-drop       & P2   & 1-arm / jaw        & --  / 400 & 8.4  \\
T15 & slide-catch             & P2   & 1-arm / jaw        & 200 / 200 & 9.6  \\
T16 & ramp-sort               & P2   & 1-arm / jaw        & 200 / 200 & 16.2 \\
T17 & domino-single           & P3   & 1-arm / jaw        & 200 / 200 & 6.3  \\
T18 & domino-select           & P3   & 1-arm / jaw        & 200 / 200 & 8.1  \\
T19 & seesaw-balance          & P3   & 1-arm / jaw        & 200 / 200 & 11.5 \\
T20 & indirect-push           & P3   & 1-arm / jaw        & 200 / 200 & 9.4  \\
T21 & gap-funnel              & P3   & 1-arm / jaw        & 200 / 200 & 10.7 \\
T22 & ball-cage-escape        & P3   & 1-arm / jaw        & 200 / 200 & 13.9 \\
T23 & tool-retrieve           & P4   & 1-arm / jaw        & 200 / 200 & 17.8 \\
T24 & cup-extract             & P4   & 1-arm / jaw        & 200 / 200 & 15.4 \\
T25 & mass-sort               & P5   & 1-arm / jaw        & 200 / 200 & 24.6 \\
T26 & occluder-push           & P5   & 1-arm / jaw        & 200 / 200 & 14.3 \\
T27 & slip-recovery           & P6   & 1-arm / jaw        & 200 / 200 & 11.2 \\
T28 & stack-collapse-recovery & P6   & 1-arm / jaw        & 200 / 200 & 19.5 \\
T29 & pick-place-ycb          & P6   & 1-arm / jaw        & 200 / 200 & 13.7 \\
T30 & shape-stack             & P7   & 1-arm / jaw        & 200 / 200 & 16.8 \\
T31 & balance-medium          & P7   & 1-arm / jaw        & 200 / 200 & 18.9 \\
T32 & balance-hard            & P7   & 1-arm / jaw        & 200 / 200 & 21.6 \\
T33 & packing                 & P7   & 1-arm / jaw        & 200 / 200 & 25.4 \\
M01 & mirror-pick-place       & M & 1-arm / jaw        & 200 / 200 & 8.7  \\
M02 & keyboard-typing         & M & 1-arm / jaw        & 200 / 200 & 28.3 \\
\bottomrule
\end{tabular}
\end{table}

For each task we target $200$ curated teleoperation and $200$ curated scripted trajectories. For the five tasks where teleoperation is infeasible (T10--T14), $400$ scripted trajectories per task are collected to compensate. After the three-stage curation pipeline, the released dataset contains $6{,}000$ teleoperation $+$ $8{,}000$ scripted $=14{,}000$ trajectories ($\approx 14$K total).

%
%
\newenvironment{taskentry}[1]{%
  \par\smallskip\noindent
  \begin{minipage}[t]{0.25\textwidth}%
    \vspace{0pt}
    {\fboxsep=1pt \fboxrule=0.4pt
     \fbox{\includegraphics[width=\dimexpr\linewidth-2\fboxrule\relax]{tasks/#1.png}}}%
  \end{minipage}\hfill
  \begin{minipage}[t]{0.72\textwidth}%
    \vspace{0pt}
}{%
  \end{minipage}%
  \par\smallskip%
}

\subsection{Per-category specifications}
\label{app:percat}
For each task we report: \textbf{Setup} (initial scene), \textbf{Goal} (success condition), \textbf{Intuition bottleneck} (the physical fact a model must infer), and \textbf{Canonical plan} (the Stage-2 reference sub-goal sequence). Only quantities that change how the task is reasoned about are listed; spawn jitter, control gains, and episode budgets are deferred to the task configs. A representative initial-state image of each scene appears to the left of its description.

\subsubsection{P1 --- Geometry-constrained grasping}

\begin{taskentry}{pb-pr-edge-slide-v1}
\noindent\textbf{T01 \texttt{edge-slide}.}~%
\textit{Setup:} a thin rectangular plate, lies flush on the table.
\textit{Goal:} lift the plate above the table. \textit{Intuition:} a top-down jaw closure cannot close around this small thickness; the plate must first be slid to overhang the table edge to expose a graspable lip.
\textit{Canonical plan:} approach the plate, push it sideways until it overhangs the table edge, pinch the exposed lip, and lift.
\href{https://lerobot-visualize-dataset.hf.space/imbench/pb-pr-edge-slide-v1/episode_0}{T01 edge-slide}
\end{taskentry}

\begin{taskentry}{pb-pr-recover-peg-insert-v1}
\noindent\textbf{T02 \texttt{recover-peg-insert}.}~%
\textit{Setup:} a rod stands vertically; a through-slot present horizontally on a side board.
\textit{Goal:} drive the peg through $\approx 95\%$ of its length into the slot, aligned with the slot axis.
\textit{Intuition:} the slot is horizontal, the arm must grasp and align the peg to the axis of the slot. A one-time $\pm 10^\circ$ yaw perturbation, applied once the peg aligns with the slot. This forces the agent to think of correct geometric alignment and constrained insertion. \textit{Canonical plan:} tip the rod onto its side against a nearby surface, regrasp it laterally so its long axis lies along the slot axis, then push it through. Realign if the line of approach is not correct. 
\href{https://lerobot-visualize-dataset.hf.space/imbench/pb-pr-recover-peg-insert-v1/episode_0}{T02 recover-peg-insert}
\end{taskentry}

\begin{taskentry}{pb-pr-sheltered-grasp-v1}
\noindent\textbf{T03 \texttt{sheltered-grasp}.}~%
\textit{Setup:} a cube sits in either the top or bottom story of a covered two-tier rack; the side face is open but the top is occluded. A goal tray sits on the table. The observations include camera views where the object is initially not visible, requiring the agent to first explore and then rely solely on the gripper camera to locate and pick the object. 
\textit{Goal:} drop the cube in the tray.
\textit{Intuition:} only lateral approach through the open face is feasible; the gripper must yaw to enter the rack.
\textit{Canonical plan:} Explore the scene to locate object, yaw the gripper to face the rack's open side, enter laterally, grasp the cube, withdraw, and drop it in the tray.
\href{https://lerobot-visualize-dataset.hf.space/imbench/pb-pr-sheltered-grasp-v1/episode_1}{T03 sheltered-grasp}
\end{taskentry}

\begin{taskentry}{pb-hetbi-cup-inversion-v1}
\noindent\textbf{T04 \texttt{cup-inversion}.}~%
\textit{Setup:} a handle-less cup starts upside-down on the table. Grippers are heterogeneous: one parallel-jaw, one suction.
\textit{Goal:} place the cup upright inside a goal region.
\textit{Intuition:} suction couples to the closed bottom but cannot achieve upright placement alone; the jaw cannot grasp the smooth rim. A reorienting suction-to-jaw handover is required. Toppling is not a reliable option since the cup just slides out. 
\textit{Canonical plan:} attach suction to the cup's exposed (upward-facing) base, lift, rotate the cup so the opening faces up, hand it off to the jaw which grabs it by the rim, and place it inside the goal region.
\href{https://lerobot-visualize-dataset.hf.space/imbench/pb-hetbi-cup-inversion-v1/episode_0}{T04 cup-inversion}
\end{taskentry}

\begin{taskentry}{pb-hobi-plate-shelf-v1}
\noindent\textbf{T05 \texttt{plate-shelf-hobi}.}~%
\textit{Setup:} A cutlery plate is positioned next to a center-shelf slot. Both arms carry parallel-jaw grippers.
\textit{Goal:} Insert the plate on edge into the shelf slot.
\textit{Intuition:} The flat plate cannot be grasped by a single arm. A symmetric two-handed pick is also infeasible, since mid air transfer is not possible without grasping, which renders final shelf placement impossible. The plate's slight incline geometry enables an alternative approach: one robotic arm presses the plate near its circumference to tilt it, creating a graspable entry for the other arm.
\textit{Canonical Plan:} Press the plate with one arm around the corner to create an edge. Grasp the plate with the other arm and transport it to the center-shelf for insertion.
\href{https://lerobot-visualize-dataset.hf.space/imbench/pb-hobi-plate-shelf-v1/episode_0}{T05 plate-shelf-hobi}
\end{taskentry}

\begin{taskentry}{pb-hetbi-plate-shelf-v1}
\noindent\textbf{T06 \texttt{plate-shelf-hetbi}.}~%
\textit{Setup:} Similar to T05, but the two arms carry different end-effectors: one suction cup, one parallel-jaw gripper.
\textit{Goal:} Insert the plate on edge into the shelf slot.
\textit{Intuition:} The asymmetric grippers enforce a role split. The suction cup lifts the plate by its broad face, replacing the press action needed in T05. The parallel-jaw gripper then grasps the plate edge. An explicit suction-to-jaw handover is required, as the suction cup cannot reach into the shelf to perform final insertion.
\textit{Canonical Plan:} Suction arm lifts the plate by its broad face. Jaw arm grasps the plate edge and inserts it into the shelf slot.
\href{https://lerobot-visualize-dataset.hf.space/imbench/pb-hetbi-plate-shelf-v1/episode_0}{T06 plate-shelf-hetbi}
\end{taskentry}

\begin{taskentry}{pb-hetbi-bowl-table-v1}
\noindent\textbf{T07 \texttt{bowl-table}.}~%
\textit{Setup:} a bowl starts upside-down in a rack. Grippers: parallel-jaw + suction.
\textit{Goal:} place the bowl upright on the table.
\textit{Intuition:} The jaw cannot grasp the bowl since no feasible affordances are available. The suction can extract the bowl from the rack but cannot reorient it cleanly on the smooth exterior; jaw must take over on the bowl rim to invert and lower it to place it in goal position. \textit{Canonical plan:} use the suction to lift the bowl out of the rack, attach jaw to its (now exposed) rim, rotate it to upright, and set it down on the table.
\href{https://lerobot-visualize-dataset.hf.space/imbench/pb-hetbi-bowl-table-v1/episode_0}{T07 bowl-table}
\end{taskentry}

\begin{taskentry}{pb-hetbi-cracker-box-suction-v1}
\noindent\textbf{T08 \texttt{cracker-box-suction}.}~%
\textit{Setup:} A cracker box stands upright on the table. A horizontal slot fixture is within reach. Only the suction arm is active; the parallel-jaw arm is frozen.
\textit{Goal:} Insert the box into the slot with its centroid aligned to the slot interior.
\textit{Intuition:} Grasping the narrow side is inefficient and requires extensive readjustment. Suction couples reliably to the broad cardboard face, making it the preferred approach.
\textit{Canonical Plan:} Align the suction cup to the broad side of the box, transport and realign to the slot and descend uniformly into it. 
\href{https://lerobot-visualize-dataset.hf.space/imbench/pb-hetbi-cracker-box-suction-v1/episode_0}{T08 cracker-box-suction}
\end{taskentry}

\begin{taskentry}{pb-hetbi-cracker-box-jaw-v1}
\noindent\textbf{T09 \texttt{cracker-box-jaw}.}~%
\textit{Setup:} Same as T08, but only the parallel-jaw arm is active; the suction arm is frozen.
\textit{Goal:} Insert the box into the slot.
\textit{Intuition:} The jaw can apply lateral forces but cannot directly fit the box into the tight slot outline. The box must be partially inserted first, then adjusted through knocking, shifting, and aligning actions to achieve final alignment.
\textit{Canonical Plan:} Grasp the box by its narrow side (flat-side grasping is not feasible with the jaw). Partially insert the box into the slot and then adjust laterally to achieve full insertion and alignment.
\href{https://lerobot-visualize-dataset.hf.space/imbench/pb-hetbi-cracker-box-jaw-v1/episode_0}{T09 cracker-box-jaw}
\end{taskentry}

\subsubsection{P2 --- Dynamics and trajectory prediction}

\begin{taskentry}{pb-pr-cube-toss-v1}
\noindent\textbf{T10 \texttt{cube-toss}.}~%
\textit{Setup:} A gripper holds a cube; the goal zone lies beyond the arm's reach.
\textit{Goal:} The cube comes to rest inside the goal zone after release.
\textit{Intuition:} Since the goal is out of reach, the cube must be thrown with sufficient velocity to reach the zone. This requires understanding projectile motion under gravity to determine appropriate throw velocity and direction.
\textit{Canonical Plan:} Lift the cube and throw it toward the goal zone with sufficient velocity and direction to land inside.
\href{https://lerobot-visualize-dataset.hf.space/imbench/pb-pr-cube-toss-v1/episode_0}{T10 cube-toss}
\end{taskentry}

\begin{taskentry}{pb-pr-pendulum-grasp-v1}
\noindent\textbf{T11 \texttt{pendulum-grasp}.}~%
\textit{Setup:} A sphere swings on an undamped pendulum at moderate amplitude.
\textit{Goal:} Grasp the bob at the swing apex.
\textit{Intuition:} Angular velocity reaches zero only at the swing extremes. Grasp timing must align with the apex phase when the bob's velocity is zero. This requires understanding periodic motion and predicting when the bob reaches the turning point.
\textit{Canonical Plan:} Track the pendulum motion and time the grasp to coincide with the swing apex, where the bob momentarily stops before reversing direction.
\href{https://lerobot-visualize-dataset.hf.space/imbench/pb-pr-pendulum-grasp-v1/episode_0}{T11 pendulum-grasp}
\end{taskentry}

\begin{taskentry}{pb-pr-linear-intercept-v1}
\noindent\textbf{T12 \texttt{linear-intercept}.}~%
\textit{Setup:} A ball rolls across the table at constant velocity, starting near one edge and heading toward the table center.
\textit{Goal:} Grasp the ball and hold it above the table for 30 consecutive steps.
\textit{Intuition:} Linear extrapolation of the constant-velocity trajectory determines the future position and timing of the ball. The gripper must be positioned at the interception point to grasp the ball when it arrives.
\textit{Canonical Plan:} Predict the ball's trajectory using its current velocity. Position the gripper at the future interception point and close the gripper as the ball arrives. Lift and hold the grasped ball above the table.
\href{https://lerobot-visualize-dataset.hf.space/imbench/pb-pr-linear-intercept-v1/episode_0}{T12 linear-intercept}
\end{taskentry}

\begin{taskentry}{pb-pr-bounce-intercept-v1}
\noindent\textbf{T13 \texttt{bounce-intercept}.}~%
\textit{Setup:} A ball is launched at a U-shaped wall enclosure and ricochets at least once before reaching the catchable zone.
\textit{Goal:} Grasp the ball and hold it above the table for 30 consecutive steps.
\textit{Intuition:} Predicting the post-bounce trajectory requires understanding how the ball's pre-bounce heading and the wall geometry jointly determine the final direction.
\textit{Canonical Plan:} Predict the ball's trajectory after bouncing off the wall using its initial heading and the wall normals. Position the gripper at the predicted interception point. Close the gripper as the ball arrives and lift it.
\href{https://lerobot-visualize-dataset.hf.space/imbench/pb-pr-bounce-intercept-v1/episode_0}{T13 bounce-intercept}
\end{taskentry}

\begin{taskentry}{pb-pr-galileo-ramp-drop-v1}
\noindent\textbf{T14 \texttt{galileo-ramp-drop}.}~%
\textit{Setup:} A steel ball sits in an open container next to a frictionless symmetric ramp.
\textit{Goal:} Release the ball on the ramp so it arrives at a goal patch on the far side at near-zero velocity.
\textit{Intuition:} Energy conservation on the frictionless ramp determines the release height needed to achieve near-zero terminal velocity at the goal patch. The ball's potential energy at release must equal its kinetic energy at the patch to minimize arrival speed.
\textit{Canonical Plan:} Approximate the required release height using energy conservation and the ramp geometry. Bring the ball at that height and release it.
\href{https://lerobot-visualize-dataset.hf.space/imbench/pb-pr-galileo-ramp-drop-v1/episode_0}{T14 galileo-ramp-drop}
\end{taskentry}

\begin{taskentry}{pb-pr-slide-catch-v1}
\noindent\textbf{T15 \texttt{slide-catch}.}~%
\textit{Setup:} An incline tilts from $10^\circ$ to $55^\circ$. Cubes with different friction coefficients sit on the incline. A graspable bin is on the table.
\textit{Goal:} Catch all cubes in the bin.
\textit{Intuition:} Each cube slides when the incline angle exceeds its friction limit, causing them to slide off at different times. The cubes closer to the bottom slide first. The bin must be positioned at the hinge edge at the right time to catch each cube in sequence.
\textit{Canonical Plan:}  Position the bin to catch the first (lowest) cube as it slides off. After catching, reposition the bin for the next cube, repeating until all are captured.
\href{https://lerobot-visualize-dataset.hf.space/imbench/pb-pr-slide-catch-v1/episode_0}{T15 slide-catch}
\end{taskentry}

\begin{taskentry}{pb-puzzle-ramp-sort-v1}
\noindent\textbf{T16 \texttt{ramp-sort}.}~%
\textit{Setup:} Multiple colored balls spawn in a low-walled box. A tilted ramp routes them downward into two bins. Barriers for goal are invisible to balls but solid to the arm.
\textit{Goal:} Every ball lands in its correct bin based on color.
\textit{Intuition:} Color-dependent rolling trajectories separate red balls from blue balls down the ramp. Any misdirected placement is irreversible. Release pose and timing determine the ball's trajectory down the ramp, so they must be chosen carefully to reach the correct bin.
\textit{Canonical Plan:} For each ball, determine the correct target bin based on its color. Release the ball at a pose such that it rolls down the ramp and lands in the correct bin.
\href{https://lerobot-visualize-dataset.hf.space/imbench/pb-puzzle-ramp-sort-v1/episode_0}{T16 ramp-sort}
\end{taskentry}

\subsubsection{P3 --- Causal and indirect action}

\begin{taskentry}{pb-pr-domino-single-v1}
\noindent\textbf{T17 \texttt{domino-single}.}~%
\textit{Setup:} a chain of dominoes leads to a red target slab.
\textit{Goal:} the cascade displaces the target by at least $5$~cm, \emph{and the gripper never touches the target}.
\textit{Intuition:} cascade physics --- toppling the first domino must propagate through the chain. Touching the target directly is an immediate, latched failure.
\textit{Canonical plan:} approach the first domino in the chain, push it toward its neighbour, and wait for the cascade to reach the target.
\href{https://lerobot-visualize-dataset.hf.space/imbench/pb-pr-domino-single-v1/episode_1}{T17 domino-single}
\end{taskentry}

\begin{taskentry}{pb-pr-domino-select-v1}
\noindent\textbf{T18 \texttt{domino-select}.}~%
\textit{Setup:} Five domino chains are present. Only one cascades fully to the target. Others contain failures: gap (domino offset perpendicular), drift (spacing increases along the chain), or the chain points to the wrong target.
\textit{Goal:} Displace the target domino via chain reaction.
\textit{Intuition:} The correct chain must be identified by visual inspection before triggering. Determine whether the chosen cascade will successfully reach the target.
\textit{Canonical Plan:} Select the structurally intact chain. Knock down the first domino in that chain.
\href{https://lerobot-visualize-dataset.hf.space/imbench/pb-pr-domino-select-v1/episode_0}{T18 domino-select}
\end{taskentry}

\begin{taskentry}{pb-pr-seesaw-balance-v1}
\noindent\textbf{T19 \texttt{seesaw-balance}.}~%
\textit{Setup:} A block sits at a random offset on one arm of a pivoted seesaw. A counterweight with unknown mass starts off the seesaw. Both masses are randomized.
\textit{Goal:} Hold the seesaw tilted within $5^\circ$ of level. 
\textit{Intuition:} Torque balance requires $m_\text{target} \cdot r_\text{target} = m_\text{counter} \cdot r_\text{counter}$. Since masses are unknown, the counterweight position must be found through trial and adjustment by observing the seesaw's tilt.
\textit{Canonical Plan:} Place the counterweight at an estimated offset. Observe the seesaw's tilt and adjust the counterweight position accordingly. Iterate until balance is achieved. 
\href{https://lerobot-visualize-dataset.hf.space/imbench/pb-pr-seesaw-balance-v1/episode_0}{T19 seesaw-balance}
\end{taskentry}

\begin{taskentry}{pb-puzzle-indirect-push-v1}
\noindent\textbf{T20 \texttt{indirect-push}.}~%
\textit{Setup:} A scatter of colored blocks and a red target block sit on the table. The goal zone is beyond the arm's direct reach.
\textit{Goal:} Drive the target into the goal zone.
\textit{Intuition:} Direct contact with the target is not possible from the arm's reach. The scattered blocks must be arranged into a contiguous chain to transmit force. Pushing the chain propagates the impulse through to the target.
\textit{Canonical Plan:} Arrange the colored blocks into a contiguous chain connecting the target to the goal zone. Push the first block in the chain so the impulse propagates through to the target, driving it into the goal zone.
\href{https://lerobot-visualize-dataset.hf.space/imbench/pb-puzzle-indirect-push-v1/episode_0}{T20 indirect-push}
\end{taskentry}

\begin{taskentry}{pb-puzzle-gap-funnel-v1}
\noindent\textbf{T21 \texttt{gap-funnel}.}~%
\textit{Setup:} A static wall with a single narrow gap runs across the table. A medium sphere is positioned in front of the wall.
\textit{Goal:} Push the sphere through the gap into the goal zone (outside robot's workspace).
\textit{Intuition:} The sphere must first be steered to align with the gap, then pushed through with appropriate velocity and direction. Path planning must account for the barrier and the narrow passage.
\textit{Canonical Plan:} Align the sphere with the gap by pushing it laterally into position. Once aligned, push the sphere through the gap with controlled velocity to reach the goal zone.
\href{https://lerobot-visualize-dataset.hf.space/imbench/pb-puzzle-gap-funnel-v1/episode_0}{T21 gap-funnel}
\end{taskentry}

\begin{taskentry}{pb-puzzle-ball-cage-escape-v1}
\noindent\textbf{T22 \texttt{ball-cage-escape}.}~%
\textit{Setup:} A thin rectangular cage containing a ball sits on a narrow stool. The cage's bottom has a hole that is blocked by the stool.
\textit{Goal:} Extract the ball from the cage onto the table.
\textit{Intuition:} Lifting and reorienting the cage will allow the ball to escape through the hole. The reorientation must account for the ball's position inside the cage to ensure it falls out.
\textit{Canonical Plan:} Lift the cage off the stool. Tilt and reorient it to align the hole with the ball's position. Continue until the ball falls out onto the table.
\href{https://lerobot-visualize-dataset.hf.space/imbench/pb-puzzle-ball-cage-escape-v1/episode_0}{T22 ball-cage-escape}
\end{taskentry}

\subsubsection{P4 --- Tool use and augmented reach}

\begin{taskentry}{pb-pr-tool-retrieve-v1}
\noindent\textbf{T23 \texttt{tool-retrieve}.}~%
\textit{Setup:} an L-shaped hook is reachable; a cube target sits beyond the arm's normal reach.
\textit{Goal:} lift the cube above the table.
\textit{Intuition:} the tool extends reach, but the gripper cannot hold both tool and target : the tool must be released \emph{before} the final grasp; releasing it after grasping the target is kinematically infeasible.
\textit{Canonical plan:} pick up the hook, drag the cube inward until it sits within normal reach, set the hook down, then grasp the cube and lift.
\href{https://lerobot-visualize-dataset.hf.space/imbench/pb-pr-tool-retrieve-v1/episode_0}{T23 tool-retrieve}
\end{taskentry}

\begin{taskentry}{pb-pr-cup-extract-v1}
\noindent\textbf{T24 \texttt{cup-extract}.}~%
\textit{Setup:} A small cuboid is trapped inside an open-top cup. The handle points toward the robot. A goal region sits away from the cup.
\textit{Goal:} Extract the cuboid from the cup and place it in the goal region.
\textit{Intuition:} The cuboid cannot be grasped directly. Only the cup handle affords a clean grasp. The cup must be manipulated by its handle to dump the cuboid out.
\textit{Canonical Plan:} Grasp the cup by its handle. Tip and invert the cup to pour the cuboid out onto the table. Move the cuboid into the goal region.
\href{https://lerobot-visualize-dataset.hf.space/imbench/pb-pr-cup-extract-v1/episode_0}{T24 cup-extract}
\end{taskentry}

\subsubsection{P5 --- Hidden state and active discovery}

\begin{taskentry}{pb-pr-mass-sort-v1}
\noindent\textbf{T25 \texttt{mass-sort}.}~%
\textit{Setup:} Three visually identical cubes have hidden masses of 50~g, 100~g, and 200~g in randomized order. Three labeled zones on the table accept the cubes in light-to-heavy order from left to right.
\textit{Goal:} Place each cube in its correct zone according to mass.
\textit{Intuition:} Mass cannot be determined visually. It is observable only through wrist force and torque during a probing lift. Each cube must be probed before placement. Any wrong-zone placement is an immediate failure.
\textit{Canonical Plan:} Lift and probe each cube to estimate its mass. Sort the cubes by inferred mass. Place them in the zones in light-to-heavy order from left to right.
\href{https://lerobot-visualize-dataset.hf.space/imbench/pb-pr-mass-sort-v1/episode_0}{T25 mass-sort}
\end{taskentry}

\begin{taskentry}{pb-pr-occluder-push-v1}
\noindent\textbf{T26 \texttt{occluder-push}.}~%
\textit{Setup:} One red target cube and two distractor cubes are each covered by a lightweight lid. A green goal patch sits elsewhere on the table.
\textit{Goal:} Place the red target cube on the goal patch.
\textit{Intuition:} The red cube cannot be identified until its lid is lifted or pushed aside. The agent must uncover each cube to locate the target.
\textit{Canonical Plan:} Lift or push aside the lids to reveal each cube. Identify the red target. Grasp the red cube and place it on the goal patch.
\href{https://lerobot-visualize-dataset.hf.space/imbench/pb-pr-occluder-push-v1/episode_0}{T26 occluder-push}
\end{taskentry}

\subsubsection{P6 --- Reactive replanning}

\begin{taskentry}{pb-pr-slip-recovery-v1}
\noindent\textbf{T27 \texttt{slip-recovery}.}~%
\textit{Setup:} A YCB object is placed on the table. The agent must pick and place it. Once during transport, the gripper unexpectedly opens for a brief window creating a synthetic failure. Sliding the object is not allowed. This tests state awareness and recovery behavior.
\textit{Goal:} Place the object inside the goal region. 
\textit{Intuition:}  The agent must detect the slip, re-grasp the object, and continue transport. 
\textit{Canonical Plan:} Lift the object and transport it toward the goal. When the gripper opens, detect the slip. Re-grasp the object and resume task.
\href{https://lerobot-visualize-dataset.hf.space/imbench/pb-pr-slip-recovery-v1/episode_0}{T27 slip-recovery}
\end{taskentry}

\begin{taskentry}{pb-puzzle-stack-collapse-recovery-v1}
\noindent\textbf{T28 \texttt{stack-collapse-recovery}.}~%
\textit{Setup:} Three cubes of different sizes start pre-stacked.
\textit{Goal:} Rebuild a stable stack.
\textit{Intuition:} An external impulse knocks over the stack, dislodging either the top cube alone or both upper cubes in different directions. The agent must rebuild a stable stack, respecting the cube sizes and mass distribution to prevent tipping.
\textit{Canonical Plan:} Detect the collapse and gather the fallen cubes. Rebuild the stack ensuring stability.
\href{https://lerobot-visualize-dataset.hf.space/imbench/pb-puzzle-stack-collapse-recovery-v1/episode_0}{T28 stack-collapse-recovery}
\end{taskentry}

\begin{taskentry}{pb-puzzle-pick-place-ycb-v1}
\noindent\textbf{T29 \texttt{pick-place-ycb}.}~%
\textit{Setup:} A single YCB object and a green goal patch sit on the table.
\textit{Goal:} Place the object at rest in the goal patch (after lifting it at least once).
\textit{Intuition:} When the lifted object approaches the patch, the patch teleports to a new location on the table. After 1--3 relocations, the patch locks in place. The agent must replan and approach the locked patch.
\textit{Canonical Plan:} Lift the object and move it toward the goal patch. When the patch relocates, track its new position, replans the path, and approach again. Repeat until the patch locks, then complete the placement.
\href{https://lerobot-visualize-dataset.hf.space/imbench/pb-puzzle-pick-place-ycb-v1/episode_0}{T29 pick-place-ycb}
\end{taskentry}

\subsubsection{P7 --- Stability and equilibrium}

\begin{taskentry}{pb-puzzle-shape-stack-v1}
\noindent\textbf{T30 \texttt{shape-stack}.}~%
\textit{Setup:} A random subset of 3--5 mixed-shape objects on the table: cylinders, cubes, prisms, hex prisms, and cone frustums.
\textit{Goal:} Stack all objects so exactly one touches the table.
\textit{Intuition:} Stable stacking requires each upper piece's center of mass to lie within the support polygon of the base. Shape-dependent geometry determines which orientations are stable.
\textit{Canonical Plan:} Select the largest or most stable object as the base. Stack objects in order, ensuring each piece's center of mass aligns over the supporting base.
\href{https://lerobot-visualize-dataset.hf.space/imbench/pb-puzzle-shape-stack-v1/episode_0}{T30 shape-stack}
\end{taskentry}

\begin{taskentry}{pb-hard-balance-medium-v1}
\noindent\textbf{T31 \texttt{balance-medium}.}~%
\textit{Setup:} A rod must be balanced on a narrow ridge of width 12~mm.
\textit{Goal:} The rod stands within $10^\circ$ of vertical and remains in contact with the ridge.
\textit{Intuition:} The 12~mm ridge provides modest tolerance for initial tilt. The rod must be released with low angular velocity and its center of mass positioned above the ridge for stable balance.
\textit{Canonical Plan:} Position the rod on the ridge with minimal tilt and angular velocity. Release the rod and allow it to balance on its own.
\href{https://lerobot-visualize-dataset.hf.space/imbench/pb-hard-balance-medium-v1/episode_0}{T31 balance-medium}
\end{taskentry}

\begin{taskentry}{pb-hard-balance-hard-v1}
\noindent\textbf{T32 \texttt{balance-hard}.}~%
\textit{Setup:} A rod must be balanced on a narrow ridge of width 8~mm.
\textit{Goal:} The rod stands within $10^\circ$ of vertical and remains in contact with the ridge.
\textit{Intuition:} The 8~mm ridge provides minimal tolerance. The rod must be released with very low residual angular velocity and its center of mass nearly perfectly positioned above the ridge for stable balance.
\textit{Canonical Plan:} Carefully position the rod on the ridge with minimal tilt and angular velocity. Release the rod and allow it to balance on its own.
\href{https://lerobot-visualize-dataset.hf.space/imbench/pb-hard-balance-hard-v1/episode_0}{T32 balance-hard}
\end{taskentry}

\begin{taskentry}{pb-pr-packing-v1}
\noindent\textbf{T33 \texttt{packing}.}~%
\textit{Setup:} A rectangular bin and a set of L-shaped pieces obtained from guillotine cuts of a rectangle. A suction gripper is used.
\textit{Goal:} Pack every piece flat inside the bin with zero gaps.
\textit{Intuition:} The L-pieces originate from a guillotine partition of the bin's footprint, so a perfect zero-gap packing exists. The agent must recover the partition structure and orient each piece to fit its slot exactly.
\textit{Canonical Plan:} Determine how the L-pieces fit together to reconstruct the original rectangular partition. Orient and place each piece into its corresponding slot in the bin, working toward a complete zero-gap packing.
\href{https://lerobot-visualize-dataset.hf.space/imbench/pb-pr-packing-v1/episode_0}{T33 packing}
\end{taskentry}

\subsubsection{Miscellaneous}

\begin{taskentry}{pb-pr-mirror-pick-place-v1}
\noindent\textbf{M01 \texttt{mirror-pick-place}.}~%
\textit{Setup:} A single cube sits on the table. Every camera frame is rotated $180^\circ$ each step.
\textit{Goal:} Lift the cube to a small height above the table.
\textit{Intuition:} The agent must distinguish between the observation frame (rotated) and the action frame (fixed). Direct imitation would invert the action direction.
\textit{Canonical Plan:} Track the cube's position in the rotated frame. Execute lift actions in the correct action frame, accounting for the $180^\circ$ rotation.
\href{https://lerobot-visualize-dataset.hf.space/imbench/pb-pr-mirror-pick-place-v1/episode_0}{M01 mirror}
\end{taskentry}

\begin{taskentry}{pb-hobi-keyboard-typing-v1}
\noindent\textbf{M02 \texttt{keyboard-typing}.}~%
\textit{Setup:} Two robotic arms face a QWERTY virtual keyboard. Arm 0 is scripted to type a subset of letters. Arm 1 is the agent. The letter partition between arms is randomized each episode.
\textit{Goal:} Type the target word in correct sequence, with each letter pressed and briefly held.
\textit{Intuition:} The agent must track which letters remain untyped and which it owns. Visual feedback on the keyboard indicates typed (bold) versus untyped (faint) letters.
\textit{Canonical Plan:} Observe which letters are marked as untyped. Identify your assigned letters from the remaining set. Press and hold each of your letters in the order required by the target word.
\href{https://lerobot-visualize-dataset.hf.space/imbench/pb-hobi-keyboard-typing-v1/episode_0}{M02 keyboard-typing}
\end{taskentry}

\section{Data Collection Infrastructure}
\label{app:data}

\subsection{Teleoperation hardware and protocol}
\label{app:teleop}
\paragraph{Single-arm.} Xbox-compatible gamepad, mapped to 6-DoF end-effector delta commands in the robot base frame. Translation magnitude 0.3 per tick, rotation magnitude 0.2 per tick. Gripper value is ramped via analog trigger; a snap-toggle to fully open/close is available on the A button with a boolean latch. \textbf{Bimanual tasks:} Same controller in arm-switching configuration: A pilot study with parallel two-handed control yielded poorer trajectory quality.

\paragraph{Filtering.} Actions are smoothed via an Exponential Moving Average (EMA) filter with $\alpha$ = 0.8, applied independently to each of the 6 motion channels (dx, dy, dz, droll, dpitch, dyaw) per arm. A per-step delta clamp of ±0.15 is applied after smoothing to limit abrupt transitions. Gripper channels pass through unfiltered to preserve snap-toggle responsiveness. Filter parameters are fixed across operators.

\paragraph{VR comparison.}
We conducted a pilot study using \textsc{Meta Quest}-based teleoperation on six representative tasks across 120 episodes. The mean post-filter success rate was $11.2\%$, compared to $58.7\%$ achieved using an XBox controller on the same tasks. We attribute this difference to the smoother and more stable motion control provided by the XBox controller, particularly in the motion regimes dominant in \textsc{IMBench}. Based on these initial results, we used the XBox controller for full data collection, while retaining support for VR teleoperation within the benchmark.

\subsection{Operator training, demographics, and consent}
Operators completed 50-60 training trajectories on each of the tasks before contributing. Calibration is repeated at the start of each session. We collected from $N{=}10$ operators (10 male; ages 21-34; all university-affiliated graduate students or research staff with backgrounds in STEM). 

\subsection{Scripted oracle authoring}
\label{app:oracles}
Each task has a hand-written policy. Oracles consume privileged simulator state and produce 6-DoF end-effector delta pose at, using the same action interface a deployed policy uses. Oracle actions are passed through the same filtering as teleop. Oracles are not motion planners; they encode task-specific strategies. For example, the \texttt{edge-slide} oracle:

\begin{verbatim}
def step(state):
    if not plate_at_edge(state):
        return slide_action(plate_pos, nearest_edge)
    elif not gripper_under_overhang(state):
        return descend_action(plate_overhang_pose)
    else:
        return close_and_lift_action()
\end{verbatim}

\subsection{Data Filtration}

\subsubsection{Human Annotation Dashboard}
Prior to automated analysis, each recorded robotic episode is reviewed through a human annotation 
dashboard. A human annotator inspects the episode( comprising RGB-D frames, joint states, 
end-effector poses, and object positions and labels scene-level metadata such as task 
completion intent, object identities, and any immediately apparent recording artifacts) and assigns a quality rating on a scale of 1 to 5. 
Only episodes receiving a rating greater than or equal to 4 are passed to the automated filtration stage; 
all others are discarded immediately. This initial screening ensures that only demonstrations 
of sufficient quality enter the downstream pipeline.

\subsubsection{VLM-Based Automated Filtering}
Following human annotation, each episode is passed through an autonomous VLM-guided filtration 
agent built on a LangChain/LangGraph tool-calling framework. The agent 
executes a deterministic preflight sequence over every episode, invoking tools for metadata and 
field integrity validation, full-episode sensor signal reading, rule-based anomaly evaluation, 
signal-processing metric computation, and FFmpeg-based visual glitch detection via blur 
($\text{Sobel gradient} < 0.3$) and scene-change ($\text{MAD score} > 10.0$) filters. Following 
this fixed pass, the agent enters a dynamic reasoning loop, issuing targeted follow-up tool 
calls such as high-FPS local frame inspection and per-channel signal plots until a 
verification gate confirms that completion status, anomaly evidence, sensor coverage, and 
signal-processing analyses are all satisfied. The agent produces a structured 
\texttt{EpisodeAnalysisReport} for each episode, flagging issues such as state discontinuities, 
action--response inconsistencies, index--timestamp mismatches, and visual anomalies, each with 
an associated severity and human-readable explanation.

\subsubsection{Human Verification Dashboard}
The agent-generated reports are surfaced through a second human verification dashboard, where 
annotators review the flagged episodes and their associated evidence including anomaly 
timestamps, sensor signal plots, and high-FPS frame windows before issuing a final 
accept or reject decision. This stage serves as a quality gate over the automated flagging, 
allowing annotators to override false positives and confirm true failures. Only episodes that 
clear this final human review are retained for training. Full prompt templates, tool schemas, 
verification gate logic, and filtration thresholds are provided in the supplementary material.

\subsection{Modality schemas}
\label{app:modalities}
\begin{table}[h]
\centering
\small
\caption{Per-trajectory streams written by the \imbench LeRobot recorder. Rates are configurable per category via \texttt{recording.save\_freq.*}; values shown are typical defaults for a 100\,Hz control loop. Cameras are stored as MP4 with per-row \texttt{frame\_index.*} pointers in Parquet.}
\label{tab:modalities}
\begin{tabular}{lllp{0.30\textwidth}}
\toprule
Stream & Shape per frame & Rate & Notes \\
\midrule
\multicolumn{4}{l}{\emph{Cameras (RGB, MP4-encoded)}}\\
\texttt{rgb\_exo\_right\_cam}      & $(480,640,3)$ uint8 & 25~Hz  & 3rd-person right (\texttt{agentview}) \\
\texttt{rgb\_exo\_left\_cam}       & $(480,640,3)$ uint8 & 25~Hz  & 3rd-person left (\texttt{sideview}) \\
\texttt{rgb\_ego\_cam}             & $(480,640,3)$ uint8 & 25~Hz  & robot-view (\texttt{robot0\_robotview}) \\
\texttt{rgb\_gripper\_cam}         & $(480,640,3)$ uint8 & 25~Hz  & wrist-mounted (\texttt{robot0\_eye\_in\_hand}) \\
\midrule
\multicolumn{4}{l}{\emph{Proprioception}}\\
\texttt{robot0\_eef\_pos}          & $(3,)$ float32       & 100~Hz & EEF position, world frame, m \\
\texttt{robot0\_eef\_quat}         & $(4,)$ float32       & 100~Hz & EEF orientation $[x,y,z,w]$ \\
\texttt{robot0\_joint\_pos}        & $(7,)$ float32       & 100~Hz & arm joints, rad \\
\texttt{robot0\_joint\_vel}        & $(7,)$ float32       & 100~Hz & arm joints, rad/s \\
\texttt{robot0\_gripper\_qpos}     & $(2,)$ float32       & 100~Hz & jaw positions \\
\midrule
\multicolumn{4}{l}{\emph{Contact \& tactile}}\\
\texttt{force\_torque}             & $(6,)$ float32       & 50~Hz  & wrist F/T in EEF frame, $\mathrm{N}$ + $\mathrm{N}{\cdot}\mathrm{m}$ \\
\texttt{force\_torque\_world}      & $(3,)$ float32       & 50~Hz  & wrist force in world frame, gravity-aligned \\
\texttt{robot0\_tactile\_left}     & $(8,8)$ float32      & 50~Hz  & left finger taxel grid \\
\texttt{robot0\_tactile\_right}    & $(8,8)$ float32      & 50~Hz  & right finger taxel grid \\
\midrule
\multicolumn{4}{l}{\emph{Action \& reward}}\\
\texttt{action}                    & $(7,)$ float32       & 100~Hz & OSC \texttt{cartesian\_delta}: $\Delta\text{pos}(3) + \Delta\boldsymbol{\omega}(3) + \text{grip}$ \\
\texttt{reward}                    & scalar float32       & 100~Hz & Sparse reward \\
\bottomrule
\end{tabular}
\end{table}

All streams are aligned to simulator wall-clock at ingestion; camera intrinsics and extrinsics are stored under \texttt{meta/cameras}.

\section{Asset Library}
\label{app:assets}

\subsection{Custom MJCF assets}
The following assets ship under IMBench:
\begin{itemize}
\item \textbf{Galileo ramp} (parametric angle $10^\circ\!-\!60^\circ$, $\mu \in [0.10, 0.50]$).
\item \textbf{Domino strips} (configurable inter-tile gap $0.6\times\!-\!1.4\times$ tile width; tile mass $0.02\!-\!0.08$~kg).
\item \textbf{Seesaw} (pivot offset $\pm 0.10$~m; beam mass $0.20\!-\!0.80$~kg).
\item \textbf{Hook tools}: J-shape (two variants) and T-shape, parametric handle length and hook radius.
\item \textbf{Handled cup} (rigid handle, variable curvature).
\item \textbf{Holed cuboid cage} (one circular hole on a configurable face).
\item \textbf{Cracker box} (4 size variants, low-friction surface).
\item \textbf{Plates} (3 thicknesses: 2, 3, 5~mm; 4 diameters).
\item \textbf{Bowl} (2 sizes).
\item \textbf{Virtual QWERTY keyboard} with per-key contact triggers and visual feedback.
\end{itemize}
All custom assets are authored in MJCF, rendered with the default \texttt{robosuite} material set, and licensed under CC-BY-4.0. Source meshes are provided in STL formats. We use the YCB-Object meshes, with collision geometry simplified by V-HACD.

\subsection{YCB usage}
\begin{table}[h]
\centering
\small
\caption{YCB object subsets per task.}
\label{tab:ycb}
\begin{tabular}{lll}
\toprule
YCB ID & Object & Tasks \\
\midrule
002 & master chef can       & pick-place-ycb \\
003 & cracker box           & sheltered-grasp \\
004 & sugar box             & pick-place-ycb \\
005 & tomato soup can       & pick-place-ycb \\
008 & pudding box           & sheltered-grasp \\
010 & banana                & pick-place-ycb \\
\bottomrule
\end{tabular}
\end{table}

\section{Evaluation}
\label{app:evalproto}

\subsection{Seeds and rollouts}
Each task ships with $S{=}20$ evaluation seeds drawn i.i.d.\ from the randomization distribution and held out from all demonstration data. Each policy is evaluated with $K{=}5$ rollout repeats per seed (different action-sampling noise but identical reset). Reported success rate is
\[
\mathrm{SR} \;=\; \frac{1}{SK}\sum_{i=1}^{S}\sum_{j=1}^{K}\mathbf{1}[s_{ij}],
\]

\subsection{Compute}
Training and evaluation were performed on AWS multi-GPU cloud instances
using NVIDIA H100 and L40S GPUs. In total, the project consumed
approximately 454 H100 GPU-hours and 1358 L40S GPU-hours across policy
training, ablations, and evaluation. Training runs used distributed
mixed-precision (BF16) execution. Interactive VLM evaluation additionally
incurred substantial API costs, with Stage~3 closed-loop evaluation costing
approximately \$1800 in total.

\section{VLM Evaluation Details}
\label{sec:vlm_evaluation}
 
The vision language model (VLM) evaluation pipeline comprises two sequentially executed stages, each independently assessed. Stage~1 evaluates visual constraint extraction from multi-view scene frames and structured metadata, while Stage~2 assesses high-level plan generation conditioned on the output of Stage~1. Detailed prompt templates for both stages are provided in the supplementary materials.
 
\subsection{Implementation Details and Agent Design}
 
The Agent operates through a two-stage architectural design that progressively grounds task planning in physical constraints derived from visual observation and sensor data.
 
In Stage~1, the agent receives multi-view camera frames of the scene alongside task descriptions, robot configuration parameters, and sensor signals (e.g., force and tactile feedback). The agent processes this information to extract physical constraints essential for task completion, such as robot mobility characteristics, motion restrictions imposed by the scene geometry, and robot-specific constraints relevant to the task. The output of Stage~1 is a constraint memory artifact, a structured representation grounding the plan in observed physical reality rather than generic procedural knowledge. This artifact maintains both episodic memory (frame-by-frame observations from each camera view) and semantic memory (facts extracted from observations, such as object contact relationships, functional roles, and critical physical properties). The artifact explicitly documents information that cannot be inferred from visual input alone, indicating what must be confirmed through sensor feedback during execution. This explicit separation of known and unknown information ensures that Stage~2 operates with full awareness of its knowledge boundaries.
 
In Stage~2, the agent receives the Stage~1 constraint artifact along with the original scene and sensor data. The agent treats all content in the Stage~1 artifact as inviolable constraints that must guide planning. Stage~2 operates through its own reasoning loop, using the extracted constraints and evidence as episodic memory and leveraging the physical facts as semantic memory to establish a grounded understanding of the scene. The agent then reasons about causal and temporal dependencies between actions, using the Stage~1 constraints to establish a partial order over required steps. This reasoning produces an ordered sequence of sub-goals, where each sub-goal specifies: (1)~the physical state to be achieved, (2)~preconditions that must hold before execution, (3)~postconditions expected after completion, and (4)~the specific Stage~1 constraint that necessitates this step. The complete sequence of sub-goals constitutes the procedural memory of the task: not merely a collection of facts about the scene, but the actionable knowledge required to execute the task step by step. All intermediate artifacts are preserved: prompts, frames, reasoning traces, validation logs, and explicit constraint links. As multiple task instances are executed, the episodic records accumulate into generalizable semantic knowledge that can be leveraged by future planning instances.
 
\subsection{Stage~1: Constraint Extraction via ReAct}
 
Stage 1 uses a structured Chain-of-Thought (CoT) prompting strategy. The model receives a single prompt containing multi-view camera frames, task metadata, controller specification, and a prescribed five step reasoning protocol (system instructions) it must execute before producing any output. All reasoning occurs within a single LLM call and gets converted into structured output for stage 2's system instructions.
 
The reasoning process unfolds in four phases:
\begin{enumerate}
    \item \textbf{Observation:} The agent systematically examines each camera view, reviews sensor data, and ingests the task description.
    \item \textbf{Reasoning:} The agent asks itself structured questions: What objects are present in the scene? Where are they located? Are there task-specific constraints or tricks imposed by scene configuration?
    \item \textbf{Constraint Ranking:} Identified constraints are prioritized according to their impact on task success, with task-critical constraints ranked highest.
    \item \textbf{Verification:} Each constraint claim is validated against actual observed frames or metadata facts to prevent hallucinations or unsupported inferences.
\end{enumerate}
 
Should the quality check fail, the agent re-executes the stage, with each attempt logged for traceability. The final output is a well-defined constraint memory artifact documenting both episodic observations (what was seen frame-by-frame across cameras) and semantic facts (object relationships, roles, and physical properties).
 
\subsection{Stage~2: Constrained Plan Generation via ReAct}
 
For Stage 2, the input comprises the constraint artifact from Stage 1 in addition to the original scene frame and the sensor information metadata. The system views all constraints in the artifact from Stage 1 as obligatory conditions on the plan, which none of the sub-goals can violate or ignore.

Stage 2 employs a structured Chain-of-Thought prompt to determine the dependency between actions and causal order through Stage 1 constraints to produce a viable partial order for actions. Prior to generation of sub-goals, the model must first explain why the simplest possible action does not work and which constraint causes its failure. Based on such reasoning, the model produces a list of sub-goals, specifying their physical outcome, preconditions, post-conditions, and the respective constraint ID from Stage 1. The model also conducts a three-way audit of failure modes to ensure the completion of the plan – F1 (all the necessary prerequisites are included in the plan), F2 (no invalid orders among sub-goals), and F3 (no informational probe sub-goals are discarded). If the combined quality score is too low, the process is repeated up to six times using the original plan as feedback, and the best solution is chosen.

\subsection{Stage~3: Closed-Loop Execution via VLM-Driven ReAct}
\label{app:stage3}

Stage~3 realises manipulation tasks through a closed-loop \textbf{ReAct (Reasoning + Acting)} cycle. At each decision step the VLM receives the full robot and scene state, expressed as structured text and camera images, and selects one motor primitive to execute. The simulator advances until that primitive converges, the resulting state becomes the next observation, and the loop continues until the task succeeds, a budget is exhausted, or the VLM signals completion.

\subsubsection*{State Representation and Visual Context}

At every step, a \textbf{structured textual observation summary} is assembled from configurable sensor channels: end-effector pose (position in metres, quaternion, roll/pitch/yaw in degrees), gripper openness, a six-axis wrist force-torque reading in sensor and world frames, per-object names and poses with physical extents (width~$\times$depth$\times$~height in metres) resolved from MuJoCo geometry, and camera calibration metadata. This summary is paired with camera images from up to four viewpoints, including an eye-in-hand gripper camera and fixed exocentric cameras, appended as base64-encoded JPEGs.

To provide temporal awareness, a \textbf{rolling frame history} of the last $H$ camera batches (default $H = 12$) is prepended to each prompt, oldest first, so the VLM can observe scene evolution rather than a single snapshot. Each additional history batch adds approximately $1{,}800$ input tokens; the rolling window bounds per-step cost independently of episode length.

\subsubsection*{Prompt Architecture and Working Memory}

Every call uses a \textbf{slot-based prompt template} that is fully self-contained: task specification, current observation summary, frame history, and working memory are all injected fresh at each step. A fixed block of robot-specific reference documentation covers the world-frame coordinate convention, canonical height landmarks (table: $z \approx 0.82$,m; safe transit: ${\sim}1.05$,m; grasp/place: $0.82$--$0.86$,m), controller scaling, and common failure modes, giving the VLM the domain knowledge needed for physically plausible planning without fine-tuning.

Because the VLM has no implicit cross-call memory, a \textbf{sliding-window working memory} (\texttt{ReactMemory}) supplies continuity. After each step an \emph{Observation}--\emph{Agent} pair is appended (abbreviated state, primitive executed, reward delta), and a window of the last $H$ pairs is serialised into the prompt. The rolling \emph{frame history} thus provides visual continuity while \emph{text memory} provides action continuity.

\subsubsection*{Action Space and Primitive Vocabulary}

The VLM selects from eight \textbf{named motor primitives}: \textsc{MoveEEF} (world-frame position and optional orientation target), \textsc{CloseGripper}, \textsc{OpenGripper}, \textsc{Lift} (vertical offset), \textsc{Descend} (lower to target $z$ holding $x$, $y$), \textsc{AtomicAction} (raw $7$D action for $N$ ticks), \textsc{Hold}, and \textsc{Done}. \textsc{MoveEEF} runs a proportional control loop that exits on convergence (position $<5$,mm, orientation $<5^\circ$), so the VLM specifies a step budget rather than a duration. A per-robot \textbf{gripper state register} propagates the last commanded open/close state into all movement primitives, preventing mid-transport drops.

\subsubsection*{VLM Integration and Control Flow}

The system supports OpenAI, Anthropic, Google, and AWS Bedrock via a unified interface. Primitives are exposed as typed JSON schemas with \texttt{tool\_choice: required}, enforcing provider-side schema validation and a structured \texttt{reasoning}/\texttt{description}/\texttt{params} response per step. Each call is wrapped in a retry loop (default: 3 attempts); failed attempts execute a \textsc{Hold}.

The ReAct loop runs for up to $K$ steps (default $K = 24$) and exits on: (1)reward $= 1.0$ (success); (2)\textsc{Done} signal; (3)~simulator termination; or (4)~step budget exhausted. A hard simulator-tick cap (default: $10{,}000$ ticks) prevents runaway primitive loops.

\subsection{Per-task VLM Stage-1 and stage-2 results}
Per-task results for Stage~1 and Stage~2 are reported in Tables~\ref{tab:vlm-stage1} and~\ref{tab:vlm-stage2}, respectively.
\label{app:vlm-stage1}
\begin{table}[htbp]
\centering
\small
\caption{Stage-1 understanding accuracy per task (\%). A response is graded correct on a probe if it satisfies all required identifications and contains no forbidden false positives; task-level accuracy is the macro-average across applicable probes}

\label{tab:vlm-stage1}
\resizebox{\textwidth}{!}{%
\begin{tabular}{llccccc} \toprule Cat. & Task & GPT-5.4-Mini & Gemma 4 & Claude-Haiku 4.5 & Claude Sonnet 4.6 & GPT-5.5\\ \midrule P1 & edge-slide & $87.5\!\pm\!4.3$ & $80.0\!\pm\!10.0$ & $\mathbf{90.0}\!\pm\!0.0$ & $\mathbf{90.0}\!\pm\!0.0$ & $\mathbf{90.0}\!\pm\!0.0$ \\ P1 & recover-peg-insert & $43.3\!\pm\!30.9$ & $36.7\!\pm\!26.2$ & $33.3\!\pm\!24.9$ & $43.3\!\pm\!30.9$ & $\mathbf{80.0}\!\pm\!0.0$ \\ P1 & sheltered-grasp & $90.0\!\pm\!0.0$ & $90.0\!\pm\!0.0$ & $90.0\!\pm\!0.0$ & $90.0\!\pm\!0.0$ & $\mathbf{100.0}\!\pm\!0.0$ \\ P1 & cup-inversion & $64.0\!\pm\!15.0$ & $74.0\!\pm\!8.0$ & $86.0\!\pm\!4.9$ & $85.0\!\pm\!5.0$ & $\mathbf{90.0}\!\pm\!0.0$ \\ P1 & plate-shelf-hobi & $84.0\!\pm\!4.9$ & $71.7\!\pm\!6.9$ & $81.7\!\pm\!9.0$ & $\mathbf{90.0}\!\pm\!0.0$ & $\mathbf{90.0}\!\pm\!0.0$ \\ P1 & plate-shelf-hetbi & $65.7\!\pm\!11.8$ & $64.3\!\pm\!10.5$ & $76.7\!\pm\!18.9$ & $\mathbf{90.0}\!\pm\!0.0$ & $\mathbf{90.0}\!\pm\!0.0$ \\ P1 & bowl-table & $54.0\!\pm\!44.1$ & $34.0\!\pm\!41.8$ & $\mathbf{92.0}\!\pm\!4.0$ & $82.0\!\pm\!31.2$ & $90.0\!\pm\!0.0$ \\ P1 & cracker-box-suction & $4.0\!\pm\!8.0$ & $4.0\!\pm\!8.0$ & $4.0\!\pm\!8.0$ & $\mathbf{13.3}\!\pm\!9.4$ & $12.0\!\pm\!24.0$ \\ P1 & cracker-box-jaw & $4.0\!\pm\!8.0$ & $4.0\!\pm\!8.0$ & $\mathbf{14.0}\!\pm\!28.0$ & $6.0\!\pm\!12.0$ & $0.0\!\pm\!0.0$ \\ P2 & cube-toss & $70.0\!\pm\!20.0$ & $40.0\!\pm\!10.0$ & $80.0\!\pm\!20.0$ & $\mathbf{95.0}\!\pm\!5.0$ & $90.0\!\pm\!0.0$ \\ P2 & pendulum-grasp & $90.0\!\pm\!0.0$ & $90.0\!\pm\!0.0$ & $90.0\!\pm\!0.0$ & $90.0\!\pm\!0.0$ & $\mathbf{100.0}\!\pm\!0.0$ \\ P2 & linear-intercept & $\mathbf{100.0}\!\pm\!0.0$ & $95.0\!\pm\!5.0$ & $95.0\!\pm\!5.0$ & $\mathbf{100.0}\!\pm\!0.0$ & $90.0\!\pm\!0.0$ \\ P2 & bounce-intercept & $\mathbf{90.0}\!\pm\!0.0$ & $\mathbf{90.0}\!\pm\!0.0$ & $\mathbf{90.0}\!\pm\!0.0$ & $\mathbf{90.0}\!\pm\!0.0$ & $\mathbf{90.0}\!\pm\!0.0$ \\ P2 & galileo-ramp-drop & $43.3\!\pm\!9.4$ & $46.7\!\pm\!4.7$ & $65.0\!\pm\!15.0$ & $\mathbf{85.0}\!\pm\!5.0$ & $66.0\!\pm\!8.0$ \\ P2 & slide-catch & $88.0\!\pm\!14.7$ & $20.0\!\pm\!0.0$ & $89.0\!\pm\!13.6$ & $\mathbf{95.0}\!\pm\!5.0$ & $80.0\!\pm\!0.0$ \\ P2 & ramp-sort & $32.0\!\pm\!14.7$ & $40.0\!\pm\!12.6$ & $74.0\!\pm\!23.3$ & $\mathbf{90.0}\!\pm\!0.0$ & $\mathbf{90.0}\!\pm\!0.0$ \\ P2 & domino-single & $\mathbf{100.0}\!\pm\!0.0$ & $\mathbf{100.0}\!\pm\!0.0$ & $\mathbf{100.0}\!\pm\!0.0$ & $\mathbf{100.0}\!\pm\!0.0$ & $90.0\!\pm\!0.0$ \\ P3 & domino-select & $55.0\!\pm\!7.6$ & $44.0\!\pm\!12.0$ & $34.0\!\pm\!10.2$ & $70.0\!\pm\!28.3$ & $\mathbf{82.0}\!\pm\!7.5$ \\ P3 & seesaw-balance & $83.3\!\pm\!9.4$ & $66.0\!\pm\!13.6$ & $82.5\!\pm\!13.0$ & $\mathbf{90.0}\!\pm\!0.0$ & $80.0\!\pm\!0.0$ \\ P3 & indirect-push & $\mathbf{100.0}\!\pm\!0.0$ & $96.0\!\pm\!4.9$ & $97.5\!\pm\!4.3$ & $\mathbf{100.0}\!\pm\!0.0$ & $\mathbf{100.0}\!\pm\!0.0$ \\ P3 & gap-funnel & $75.0\!\pm\!5.0$ & $90.0\!\pm\!0.0$ & $\mathbf{100.0}\!\pm\!0.0$ & $\mathbf{100.0}\!\pm\!0.0$ & $\mathbf{100.0}\!\pm\!0.0$ \\ P3 & ball-cage-escape & $90.0\!\pm\!0.0$ & $90.0\!\pm\!0.0$ & $90.0\!\pm\!0.0$ & $90.0\!\pm\!0.0$ & $\mathbf{100.0}\!\pm\!0.0$ \\ P3 & tool-retrieve & $90.0\!\pm\!0.0$ & $90.0\!\pm\!0.0$ & $80.0\!\pm\!10.0$ & $90.0\!\pm\!0.0$ & $\mathbf{100.0}\!\pm\!0.0$ \\ P4 & cup-extract & $\mathbf{90.0}\!\pm\!0.0$ & $86.0\!\pm\!8.0$ & $70.0\!\pm\!21.0$ & $\mathbf{90.0}\!\pm\!0.0$ & $86.0\!\pm\!8.0$ \\ P5 & mass-sort & $\mathbf{100.0}\!\pm\!0.0$ & $90.0\!\pm\!0.0$ & $\mathbf{100.0}\!\pm\!0.0$ & $\mathbf{100.0}\!\pm\!0.0$ & $90.0\!\pm\!0.0$ \\ P5 & occluder-push & $90.0\!\pm\!0.0$ & $90.0\!\pm\!0.0$ & $67.5\!\pm\!39.0$ & $95.0\!\pm\!5.0$ & $\mathbf{98.3}\!\pm\!3.7$ \\ P6 & slip-recovery & $16.7\!\pm\!12.5$ & $6.7\!\pm\!9.4$ & $30.0\!\pm\!24.5$ & $33.3\!\pm\!24.9$ & $\mathbf{80.0}\!\pm\!0.0$ \\ P6 & stack-collapse-recovery & $\mathbf{80.0}\!\pm\!10.0$ & $75.0\!\pm\!15.0$ & $40.0\!\pm\!0.0$ & $55.0\!\pm\!15.0$ & $0.0\!\pm\!0.0$ \\ P6 & pick-place-ycb & $90.0\!\pm\!0.0$ & $70.0\!\pm\!0.0$ & $40.0\!\pm\!0.0$ & $\mathbf{100.0}\!\pm\!0.0$ & $90.0\!\pm\!10.0$ \\ P7 & shape-stack & $6.7\!\pm\!9.4$ & $23.3\!\pm\!20.5$ & $0.0\!\pm\!0.0$ & $25.0\!\pm\!25.0$ & $\mathbf{60.0}\!\pm\!11.0$ \\ P7 & balance-medium & $63.3\!\pm\!4.7$ & $43.3\!\pm\!18.9$ & $13.3\!\pm\!9.4$ & $86.7\!\pm\!4.7$ & $\mathbf{90.0}\!\pm\!0.0$ \\ P7 & balance-hard & $66.7\!\pm\!17.0$ & $33.3\!\pm\!47.1$ & $76.7\!\pm\!20.5$ & $\mathbf{93.3}\!\pm\!4.7$ & $72.0\!\pm\!9.8$ \\ P7 & packing & $50.0\!\pm\!0.0$ & $50.0\!\pm\!0.0$ & $50.0\!\pm\!0.0$ & $55.0\!\pm\!5.0$ & $\mathbf{100.0}\!\pm\!0.0$ \\ M & mirror-pick-place & $0.0\!\pm\!0.0$ & $0.0\!\pm\!0.0$ & $\mathbf{4.0}\!\pm\!8.0$ & $0.0\!\pm\!0.0$ & $0.0\!\pm\!0.0$ \\ M & keyboard-typing & $10.0\!\pm\!10.0$ & $0.0\!\pm\!0.0$ & $0.0\!\pm\!0.0$ & $0.0\!\pm\!0.0$ & $\mathbf{16.7}\!\pm\!23.6$ \\ \bottomrule \end{tabular}
}
\end{table}

\label{app:vlm-stage2}
\begin{table}[htbp]
\centering
\small
\caption{Stage-2 plan correctness per task (\%). Plans graded correct if a human verifier finds the generated plan sufficient to execute the task correctly.}
\label{tab:vlm-stage2}
\resizebox{\textwidth}{!}{%
\begin{tabular}{llccccc} \toprule Cat. & Task & GPT-5.4-Mini & Gemma 4 & Claude-Haiku 4.5 & Claude Sonnet 4.6 & GPT-5.5\\ \midrule P1 & edge-slide & $87.5\!\pm\!4.3$ & $80.0\!\pm\!10.0$ & $\mathbf{90.0}\!\pm\!0.0$ & $\mathbf{90.0}\!\pm\!0.0$ & $\mathbf{90.0}\!\pm\!0.0$ \\ P1 & recover-peg-insert & $26.7\!\pm\!18.9$ & $30.0\!\pm\!21.6$ & $26.7\!\pm\!18.9$ & $46.7\!\pm\!33.0$ & $\mathbf{80.0}\!\pm\!0.0$ \\ P1 & sheltered-grasp & $85.0\!\pm\!5.0$ & $90.0\!\pm\!0.0$ & $90.0\!\pm\!0.0$ & $90.0\!\pm\!0.0$ & $\mathbf{100.0}\!\pm\!0.0$ \\ P1 & cup-inversion & $34.0\!\pm\!41.8$ & $82.0\!\pm\!4.0$ & $88.0\!\pm\!4.0$ & $40.0\!\pm\!20.0$ & $\mathbf{90.0}\!\pm\!0.0$ \\ P1 & plate-shelf-hobi & $64.0\!\pm\!10.2$ & $46.7\!\pm\!7.5$ & $43.3\!\pm\!18.0$ & $70.0\!\pm\!8.2$ & $\mathbf{75.0}\!\pm\!8.7$ \\ P1 & plate-shelf-hetbi & $55.7\!\pm\!7.3$ & $55.7\!\pm\!14.0$ & $76.7\!\pm\!18.9$ & $85.7\!\pm\!4.9$ & $\mathbf{90.0}\!\pm\!0.0$ \\ P1 & bowl-table & $50.0\!\pm\!41.5$ & $32.0\!\pm\!39.7$ & $\mathbf{92.0}\!\pm\!4.0$ & $74.0\!\pm\!37.2$ & $88.0\!\pm\!4.0$ \\ P1 & cracker-box-suction & $4.0\!\pm\!8.0$ & $0.0\!\pm\!0.0$ & $0.0\!\pm\!0.0$ & $0.0\!\pm\!0.0$ & $\mathbf{8.0}\!\pm\!16.0$ \\ P1 & cracker-box-jaw & $0.0\!\pm\!0.0$ & $\mathbf{4.0}\!\pm\!8.0$ & $\mathbf{4.0}\!\pm\!8.0$ & $0.0\!\pm\!0.0$ & $0.0\!\pm\!0.0$ \\ P2 & cube-toss & $45.0\!\pm\!45.0$ & $15.0\!\pm\!15.0$ & $80.0\!\pm\!20.0$ & $75.0\!\pm\!15.0$ & $\mathbf{90.0}\!\pm\!0.0$ \\ P2 & pendulum-grasp & $90.0\!\pm\!0.0$ & $90.0\!\pm\!0.0$ & $90.0\!\pm\!0.0$ & $90.0\!\pm\!0.0$ & $\mathbf{100.0}\!\pm\!0.0$ \\ P2 & linear-intercept & $\mathbf{100.0}\!\pm\!0.0$ & $95.0\!\pm\!5.0$ & $95.0\!\pm\!5.0$ & $\mathbf{100.0}\!\pm\!0.0$ & $90.0\!\pm\!0.0$ \\ P2 & bounce-intercept & $76.7\!\pm\!9.4$ & $70.0\!\pm\!10.0$ & $86.7\!\pm\!4.7$ & $\mathbf{90.0}\!\pm\!0.0$ & $\mathbf{90.0}\!\pm\!0.0$ \\ P2 & galileo-ramp-drop & $26.7\!\pm\!4.7$ & $23.3\!\pm\!4.7$ & $40.0\!\pm\!10.0$ & $\mathbf{75.0}\!\pm\!15.0$ & $30.0\!\pm\!25.3$ \\ P2 & slide-catch & $0.0\!\pm\!0.0$ & $0.0\!\pm\!0.0$ & $15.0\!\pm\!21.4$ & $\mathbf{95.0}\!\pm\!5.0$ & $0.0\!\pm\!0.0$ \\ P2 & ramp-sort & $26.0\!\pm\!8.0$ & $38.0\!\pm\!14.7$ & $74.0\!\pm\!23.3$ & $\mathbf{90.0}\!\pm\!0.0$ & $\mathbf{90.0}\!\pm\!0.0$ \\ P2 & domino-single & $\mathbf{100.0}\!\pm\!0.0$ & $\mathbf{100.0}\!\pm\!0.0$ & $\mathbf{100.0}\!\pm\!0.0$ & $\mathbf{100.0}\!\pm\!0.0$ & $90.0\!\pm\!0.0$ \\ P3 & domino-select & $45.0\!\pm\!12.6$ & $34.0\!\pm\!12.0$ & $26.0\!\pm\!12.0$ & $70.0\!\pm\!28.3$ & $\mathbf{78.0}\!\pm\!14.7$ \\ P3 & seesaw-balance & $76.7\!\pm\!12.5$ & $42.0\!\pm\!18.3$ & $70.0\!\pm\!12.2$ & $\mathbf{85.0}\!\pm\!5.0$ & $80.0\!\pm\!0.0$ \\ P3 & indirect-push & $\mathbf{100.0}\!\pm\!0.0$ & $68.0\!\pm\!39.2$ & $77.5\!\pm\!33.4$ & $\mathbf{100.0}\!\pm\!0.0$ & $\mathbf{100.0}\!\pm\!0.0$ \\ P3 & gap-funnel & $90.0\!\pm\!0.0$ & $75.0\!\pm\!15.0$ & $\mathbf{100.0}\!\pm\!0.0$ & $\mathbf{100.0}\!\pm\!0.0$ & $\mathbf{100.0}\!\pm\!0.0$ \\ P3 & ball-cage-escape & $90.0\!\pm\!0.0$ & $90.0\!\pm\!0.0$ & $90.0\!\pm\!0.0$ & $90.0\!\pm\!0.0$ & $\mathbf{100.0}\!\pm\!0.0$ \\ P3 & tool-retrieve & $60.0\!\pm\!30.0$ & $90.0\!\pm\!0.0$ & $40.0\!\pm\!10.0$ & $60.0\!\pm\!30.0$ & $\mathbf{100.0}\!\pm\!0.0$ \\ P4 & cup-extract & $\mathbf{74.0}\!\pm\!10.2$ & $52.0\!\pm\!18.3$ & $24.0\!\pm\!4.9$ & $25.0\!\pm\!5.0$ & $70.0\!\pm\!11.0$ \\ P5 & mass-sort & $95.0\!\pm\!5.0$ & $90.0\!\pm\!0.0$ & $\mathbf{100.0}\!\pm\!0.0$ & $\mathbf{100.0}\!\pm\!0.0$ & $90.0\!\pm\!0.0$ \\ P5 & occluder-push & $88.0\!\pm\!4.0$ & $90.0\!\pm\!0.0$ & $67.5\!\pm\!39.0$ & $95.0\!\pm\!5.0$ & $\mathbf{98.3}\!\pm\!3.7$ \\ P6 & slip-recovery & $6.7\!\pm\!9.4$ & $23.3\!\pm\!20.5$ & $36.7\!\pm\!26.2$ & $36.7\!\pm\!26.2$ & $\mathbf{80.0}\!\pm\!0.0$ \\ P6 & stack-collapse-recovery & $55.0\!\pm\!15.0$ & $10.0\!\pm\!10.0$ & $\mathbf{60.0}\!\pm\!0.0$ & $25.0\!\pm\!25.0$ & $0.0\!\pm\!0.0$ \\ P6 & pick-place-ycb & $30.0\!\pm\!0.0$ & $20.0\!\pm\!0.0$ & $50.0\!\pm\!0.0$ & $\mathbf{90.0}\!\pm\!0.0$ & $\mathbf{90.0}\!\pm\!10.0$ \\ P7 & shape-stack & $6.7\!\pm\!9.4$ & $6.7\!\pm\!9.4$ & $0.0\!\pm\!0.0$ & $0.0\!\pm\!0.0$ & $\mathbf{60.0}\!\pm\!11.0$ \\ P7 & balance-medium & $20.0\!\pm\!28.3$ & $6.7\!\pm\!9.4$ & $6.7\!\pm\!9.4$ & $23.3\!\pm\!33.0$ & $\mathbf{88.0}\!\pm\!4.0$ \\ P7 & balance-hard & $6.7\!\pm\!9.4$ & $33.3\!\pm\!47.1$ & $56.7\!\pm\!41.9$ & $\mathbf{83.3}\!\pm\!17.0$ & $72.0\!\pm\!9.8$ \\ P7 & packing & $25.0\!\pm\!25.0$ & $0.0\!\pm\!0.0$ & $50.0\!\pm\!0.0$ & $50.0\!\pm\!0.0$ & $\mathbf{100.0}\!\pm\!0.0$ \\ M & mirror-pick-place & $0.0\!\pm\!0.0$ & $\mathbf{4.0}\!\pm\!8.0$ & $\mathbf{4.0}\!\pm\!8.0$ & $0.0\!\pm\!0.0$ & $0.0\!\pm\!0.0$ \\ M & keyboard-typing & $10.0\!\pm\!10.0$ & $0.0\!\pm\!0.0$ & $0.0\!\pm\!0.0$ & $0.0\!\pm\!0.0$ & $\mathbf{16.7}\!\pm\!23.6$ \\ \bottomrule 
\end{tabular}
}
\end{table}


\begin{table}[htbp]
\centering
\small
\caption{Stage-3 action success rate per task (\%) for GPT-5.5 without and with privileged object context.}
\label{tab:vlm-stage3-gpt55}
\begin{tabular*}{\textwidth}{l@{\extracolsep{\fill}}lcc}
\toprule
Cat. & Task & GPT-5.5 & GPT-5.5-obj \\
\midrule
P1 & edge-slide              & $0.0$ & $0.0$ \\
P1 & recover-peg-insert      & $0.0$ & $0.0$ \\
\midrule
P2 & bounce-intercept        & $0.0$ & $0.0$ \\
P2 & slide-catch             & $0.0$ & $0.0$ \\
\midrule
P3 & domino-single           & $\mathbf{100.0}$ & $\mathbf{100.0}$ \\
P3 & domino-select           & $50.0$ & $\mathbf{100.0}$ \\
\midrule
P4 & tool-retrieve           & $0.0$ & $0.0$ \\
P4 & cup-extract             & $0.0$ & $0.0$ \\
\midrule
P5 & mass-sort               & $0.0$ & $0.0$ \\
P5 & occluder-push           & $0.0$ & $0.0$ \\
\midrule
P6 & slip-recovery           & $0.0$ & $0.0$ \\
P6 & pick-place-ycb          & $0.0$ & $0.0$ \\
\midrule
P7 & shape-stack             & $0.0$ & $0.0$ \\
P7 & balance-medium          & $0.0$ & $0.0$ \\
\midrule
M  & mirror-pick-place       & $30.0$ & $\mathbf{100.0}$ \\
M  & keyboard-typing         & $0.0$ & $0.0$ \\
\bottomrule
\end{tabular*}
\end{table}

\clearpage

\section{Policy Training Details}
\label{app:training}

\subsection{Diffusion Policy}
We train Diffusion Policy (DP) from scratch using the LeRobot diffusion policy implementation. Unlike $\pi_{0.5}$ and GR00T N1.5, no pretrained policy checkpoint is finetuned. 

\begin{table}[htbp]
\centering
\small
\caption{Diffusion Policy training settings.}
\label{tab:dp-hp}
\begin{tabular}{@{}p{0.33\textwidth}p{0.62\textwidth}@{}}
\toprule
Hyperparameter & Value \\
\midrule
Trainable parameters & Trained from scratch \\
Optimizer & AdamW \\
Learning rate & $1\!\times\!10^{-4}$ \\
LR schedule & Cosine with 500-step warmup \\
Batch size & 32 \\
Training steps & 50k steps \\
Weight decay & $1\!\times\!10^{-6}$ \\
Checkpoint save frequency & every 5K steps \\
\bottomrule
\end{tabular}
\end{table}

\subsection{$\pi_{0.5}$ Finetuning}
We finetune the public \texttt{lerobot/pi05\_base} checkpoint.

\begin{table}[h]
\centering
\small
\caption{$\pi_{0.5}$ finetuning settings.}
\label{tab:pi-hp}
\begin{tabular}{@{}p{0.33\textwidth}p{0.62\textwidth}@{}}
\toprule
Hyperparameter & Value \\
\midrule
Base checkpoint & \texttt{lerobot/pi05\_base} \\
LoRA config & Rank 64, \(\alpha=128\), dropout 0.05 \\
Observations & 4 RGB cameras (front/left/right/gripper), language instruction \\
State/action dims & Native single-arm 7D action, bimanual 14D; padded to unified format \\
Optimizer & AdamW \\
Learning rate & \(3\times10^{-4}\) \\
LR schedule & Cosine decay, 200-step warmup \\
Batch size & 128 \\
Training steps &  50K \\
Gradient clip norm & 1.0 \\
Weight decay & 0.0 \\
\bottomrule
\end{tabular}
\end{table}

\subsection{GR00T N1.5 Finetuning}
We finetune \texttt{nvidia/GR00T-N1.5-3B} through LeRobot's GR00T policy stack. 

\begin{table}[h]
\centering
\small
\caption{GR00T N1.5 finetuning settings.}
\label{tab:groot-hp}
\begin{tabular}{@{}p{0.33\textwidth}p{0.62\textwidth}@{}}
\toprule
Hyperparameter & Value \\
\midrule
Base checkpoint & \texttt{nvidia/GR00T-N1.5-3B} \\
LoRA config & Rank 64, \(\alpha=128\), dropout 0.05 \\
Observations & 4 RGB cameras (resized to 224\(\times\)224), 15D state, 7D action \\
Optimizer & AdamW \\
Learning rate & \(3\times10^{-4}\) \\
LR schedule & Cosine decay, 400-step warmup \\
Batch size & 64 \\
Training steps & 20K
\\
Gradient clip norm & 1.0 \\
Weight decay & 0.0 \\
\bottomrule
\end{tabular}
\end{table}

\subsection{Software Environment}
Both $\pi_{0.5}$ and GR00T N1.5 are trained in bf16 mixed precision. Flash attention is enabled for GR00T System-1 VLM attention and the $\pi_{0.5}$ PaliGemma backbone.
\begin{table}[h]
\centering
\small
\caption{Software environment used for policy training.}
\label{tab:train-env}
\begin{tabular}{@{}p{0.33\textwidth}p{0.62\textwidth}@{}}
\toprule
Package & Version \\
\midrule
Python & 3.12 \\
PyTorch & 2.7.1 + cu126 \\
Flash-Attention & 2.7.4.post1 \\
robosuite & 1.5 \\
MuJoCo & 3.1.2 \\
CUDA & 12.6 \\
\bottomrule
\end{tabular}
\end{table}

\section{Per-task Policy Results}
\label{app:per-task-policy}

\begin{table}[htbp]
\centering
\small
\setlength{\tabcolsep}{12pt}   
\renewcommand{\arraystretch}{1.0} 
\caption{\textbf{Policy success rates (\%) per task} on \imbench. Bold marks the best per row.}
\label{tab:per-policy-results}
\begin{tabularx}{\textwidth}{@{}lXcccccc@{}}
\toprule
& & \multicolumn{2}{c}{\textbf{$\pi_{0.5}$}}
  & \multicolumn{2}{c}{\textbf{GR00T~1.5}}
  & \multicolumn{1}{c}{\textbf{DP}} \\
\cmidrule(lr){3-4}\cmidrule(lr){5-6}
& & \textbf{ZS} & \textbf{FT} & \textbf{ZS} & \textbf{FT} & \textbf{Full-training} \\
\midrule

T01 & edge-slide              & 0.00 & 0.07 & 0.00 & 0.00 & \textbf{0.28} \\
T02 & recover-peg-insert      & 0.00 & 0.03 & 0.00 & 0.00 & \textbf{0.29} \\
T03 & sheltered-grasp         & 0.00 & 0.10 & 0.00 & 0.00 & \textbf{0.30} \\
T04 & cup-inversion           & 0.00 & 0.00 & 0.00 & 0.00 & 0.00 \\
T05 & plate-shelf-hobi        & 0.00 & 0.00 & 0.00 & 0.00 & 0.00 \\
T06 & plate-shelf-hetbi       & 0.00 & 0.00 & 0.00 & 0.00 & 0.00 \\
T07 & bowl-table              & 0.00 & \textbf{0.27} & 0.00 & 0.00 & 0.00 \\
T08 & cracker-box-suction     & 0.00 & 0.00 & 0.00 & 0.00 & 0.00 \\
T09 & cracker-box-jaw         & 0.00 & 0.00 & 0.00 & 0.00 & 0.00 \\

T10 & cube-toss               & 0.00 & \textbf{0.12} & 0.00 & 0.00 & 0.09 \\
T11 & pendulum-grasp          & 0.08 & 0.28 & 0.00 & 0.08 & \textbf{0.35} \\
T12 & linear-intercept        & 0.00 & \textbf{0.04} & 0.00 & \textbf{0.04} & 0.01 \\
T13 & bounce-intercept        & 0.00 & 0.00 & 0.00 & \textbf{0.04} & 0.00 \\
T14 & galileo-ramp-drop       & 0.00 & 0.00 & 0.00 & 0.00 & \textbf{0.34} \\
T15 & slide-catch             & 0.00 & \textbf{0.10} & 0.00 & 0.00 & 0.00 \\
T16 & ramp-sort               & 0.00 & 0.00 & 0.00 & 0.00 & \textbf{0.07} \\

T17 & domino-single           & 0.28 & \textbf{0.73} & 0.08 & 0.25 & 0.70 \\
T18 & domino-select           & 0.00 & 0.32 & 0.08 & 0.17 & \textbf{0.80} \\
T19 & seesaw-balance          & 0.00 & 0.10 & 0.00 & 0.00 & \textbf{0.13} \\
T20 & indirect-push           & 0.00 & 0.07 & 0.00 & 0.00 & \textbf{0.08} \\
T21 & gap-funnel              & 0.00 & 0.06 & 0.00 & 0.00 & \textbf{0.07} \\
T22 & ball-cage-escape        & 0.22 & \textbf{0.45} & 0.00 & 0.22 & 0.43 \\

T23 & tool-retrieve           & \textbf{0.18} & 0.00 & 0.00 & 0.00 & 0.00 \\
T24 & cup-extract             & 0.00 & 0.02 & 0.00 & 0.00 & \textbf{0.07} \\

T25 & mass-sort               & 0.00 & 0.00 & 0.00 & 0.00 & 0.00 \\
T26 & occluder-push           & 0.00 & 0.15 & 0.00 & 0.00 & \textbf{0.50} \\

T27 & slip-recovery           & 0.00 & 0.22 & 0.00 & 0.00 & \textbf{0.44} \\
T28 & stack-collapse-recovery & 0.00 & 0.00 & 0.00 & 0.00 & 0.00 \\
T29 & pick-place-ycb          & 0.00 & \textbf{0.30} & 0.00 & 0.00 & 0.28 \\

T30 & shape-stack             & 0.00 & 0.00 & 0.00 & 0.00 & 0.00 \\
T31 & balance-medium          & 0.00 & \textbf{0.71} & 0.00 & 0.02 & 0.28 \\
T32 & balance-hard            & 0.00 & \textbf{0.58} & 0.00 & 0.03 & 0.20 \\
T33 & packing                 & 0.00 & 0.00 & 0.00 & 0.00 & 0.00 \\

M01 & mirror-pick-place       & 0.02 & 0.00 & 0.00 & 0.00 & \textbf{0.98} \\
M02 & keyboard-typing         & 0.00 & 0.13 & 0.00 & 0.04 & \textbf{0.32} \\

\bottomrule
\end{tabularx}
\end{table}

\section{Out-of-Distribution Specifications}
\label{app:ood}

For each task we list the physical axis the OOD evaluation probes, the training distribution along that axis, and the OOD distribution. OOD distributions are designed to require \emph{extrapolation} of the reasoning to diverse scenarios. 
\newcommand{\oodrow}[4]{%
  \noindent\textbf{#1} (#2): #3 $\;\rightarrow\;$ #4\\[3pt]
}

\oodrow{P1 sheltered-grasp}
{Geometry-aware grasping under occlusion}
{object sheltered from a canonical lateral side}
{object placed on the opposite side, requiring adaptation of grasp approach}
\oodrow{P2 pendulum-grasp}
{Dynamic interception of a swinging object}
{pendulum swings in the sagittal plane ($0^\circ$)}
{pendulum swing plane rotated to $45^\circ$, altering timing and approach trajectory}
\oodrow{P3 domino-single}
{Sequential contact-rich manipulation}
{axis-aligned domino chain configuration}
{domino chain rotated diagonally by $45^\circ$, changing spatial interaction geometry}
\oodrow{P4 tool-retrieve}
{Tool-use for indirect object interaction}
{standard hook orientation used for retrieval}
{hook mirror-flipped with the tip on the opposite side, requiring strategy adaptation}
\oodrow{P5 occluder-push}
{Manipulation under increasing clutter and occlusion}
{single occluder lid covering the target}
{2-3 additional lids stacked on top, increasing obstruction complexity}
\oodrow{P6 slip-recovery}
{Reactive replanning after unexpected disturbances}
{single slip event during object transport}
{two independent slips introduced during transport, requiring repeated recovery}
\oodrow{P7 balance-medium}
{Balancing under geometric distribution shifts}
{limited base and rod orientations during training}
{both base and rod orientations shifted outside the training distribution}
\oodrow{M keyboard-typing}
{Language-conditioned long-horizon typing}
{fixed vocabulary words seen during training}
{previously unseen words requiring compositional generalization}

\section{Detailed Analysis of Benchmark Results and Emergent Insights}
\label{app:insights}

We evaluated performance across the three evaluation stages: constraint
understanding, plan generation, and policy execution to
identify where competence is gained, lost, or never acquired. Numbers in this
section reference Tables~\ref{tab:vlm-stage1}, \ref{tab:vlm-stage2}, and \ref{tab:per-policy-results}.


\subsection{Stage 1: Physical Constraint Understanding}
\label{app:insights_stage1}

Stage~1 results show that current VLMs can identify basic physical constraints
needed for common manipulation tasks, yet consistently miss fine-grained details
that are critical for task success. Models perform relatively well on simpler
tasks: domino-single and indirect-push reached $90$--$100\%$ across all models,
linear-intercept scored $95$--$100\%$, and bounce-intercept reached $90\%$
uniformly. Performance drops considerably once tasks require precise physical
reasoning.

\textbf{Understands the goal but skips small critical details.}
Balance and stacking tasks expose this gap clearly. For balance-medium, most
models scored in the $43$--$87\%$ range (Haiku and GPT-5.5 were outliers on this), while shape-stack scores were near
floor level (GPT-5.4-Mini $6.7\%$, Claude-Haiku $0\%$, Claude-Sonnet $25\%$), except GPT-5.5 which scored $60\%$,
indicating that COM-grounded reasoning is rarely produced for most models. Interestingly,
balance-hard showed higher average accuracy than balance-medium for several
models—Claude-Haiku improved from $13.3\%$ to $76.7\%$ and Claude-Sonnet from
$86.7\%$ to $93.3\%$—because hard-balance tasks were designed to test sub-point
precision, which directly prompted COM positioning and alignment reasoning.
In insertion tasks (recover-peg-insert), scores remained moderate and variable
($33$--$80\%$), consistent with models identifying the high-level insert goal
while omitting the required pre-insertion alignment constraint.

\textbf{Recognizes the object but misses its mechanical role.}
Cracker-box tasks show the most severe failure in Stage~1. Both cracker-box-jaw
and cracker-box-suction scored near zero across all models: jaw scores ranged
from $0\%$ (GPT-5.5) to $14\%$ (Claude-Haiku), and suction scores ranged from
$4\%$ to $13.3\%$. Models identified the object and the relevant surface but never produced the specific mechanical constraints required for a valid grasp, such as gripper alignment angle for jaw grasping or correct face selection for suction.

\textbf{Understands start and end but misses the transition.}
Orientation tasks such as cup-inversion scored in the $64$--$90\%$ range, and
plate-shelf tasks (hobi/hetbi) scored $65$--$90\%$, suggesting that models
recognized the target configuration. However, they consistently omitted the
rotation or reorientation step required to reach it. Tool-retrieve scores were
high ($80$--$100\%$), meaning models knew the hook was relevant, but as
discussed in Stage~2, the hook-to-gripper transition was still missing from
their plans.

\textbf{Detects hidden constraints only when the test is direct.}
Models handled observable hidden constraints well: occluder-push scored
$90$--$98\%$ across models, and mass-sort reached $90$--$100\%$, because both
tasks require a single direct action (move the plate, lift the cube) to expose
the missing information. In contrast, mirror-pick-place scored $0\%$ for all
models except Claude-Haiku at $4\%$. This task required reasoning that the
visual input was horizontally flipped and that left-right motions must be
reversed, a non-obvious, abstract constraint that no model reliably identified.

Overall, Stage~1 failures follow a consistent pattern: models understand the
objective and identify the relevant objects but skip the specific physical
constraint that makes the task succeed, \textbf{describing tasks procedurally
rather than reasoning from physical principles.}

\subsection{Stage 2 Analysis: Constraint-to-Plan Conversion}
\label{subsec:stage2_results}

 Since Stage~2 requires grounding Stage~1 understanding into concrete actions, the relatively small performance gap between the two stages suggests that once models understand the constraints, they can usually translate them into appropriate actions. We discuss when the models struggled to convert correctly identified physical constraints from Stage~1 into an ordered and executable action plan.

\textbf{Mentioned dynamic constraints but did not track them in the plan.}
The slide-catch task illustrates this gap most sharply. Stage~1 scores were
strong for most models (GPT-5.4-Mini $88\%$, Claude-Haiku $89\%$, Claude-Sonnet
$95\%$), showing they understood the ramp-tilt dynamics and bin-timing
constraint. Yet Stage~2 scores collapsed to $0\%$ for GPT-5.4-Mini, Gemma~4,
and GPT-5.5, and only $15\%$ for Claude-Haiku, with Claude-Sonnet the only
model maintaining a high score at $95\%$. Models produced plans that named
the bin and the cube but contained no steps for tracking the tilt angle,
predicting the release point, or adjusting the bin position dynamically.
Knowing the constraint existed was not enough to generate a plan that
satisfied it.

\textbf{Knew a tool switch was needed but did not include it in the plan.}
In tool-retrieve tasks, Stage~1 scores were uniformly high ($80$--$100\%$
across all models), confirming that the hook-to-gripper transition was
identified. However, Stage~2 scores dropped noticeably for several models:
Claude-Haiku fell from $80\%$ to $40\%$ and GPT-5.4-Mini from $90\%$ to
$60\%$. Even in cases where the transition appeared in the output, plans often
described dragging the hook across the surface without releasing it, which is
an incorrect action sequence. This shows that models can recognize a
constraint in Stage~1 and still fail to place it correctly in the plan.

\textbf{Detected slip but did not plan real recovery.}
Slip-recovery was among the weakest tasks across both stages. Stage~1 scores
were already low (GPT-5.4-Mini $16.7\%$, Gemma~4 $6.7\%$, Claude-Haiku $30\%$,
Claude-Sonnet $33.3\%$), and Stage~2 scores remained similarly low or declined
further (GPT-5.4-Mini $6.7\%$, Gemma~4 $23.3\%$, Claude-Haiku $36.7\%$,
Claude-Sonnet $36.7\%$), with GPT-5.5 the only model reaching $80\%$ in
both stages. Where slip was mentioned, plans typically added only a monitor or
grip-force step. Models planned for observation but not for the corrective action.

\textbf{Recognized collapse but planned only a partial rebuild.}
Stack-collapse-recovery showed a consistent downward shift from Stage~1 to
Stage~2. GPT-5.4-Mini went from $80\%$ to $55\%$, Gemma~4 from $75\%$ to
$10\%$, and Claude-Sonnet from $55\%$ to $25\%$. GPT-5.5 scored $0\%$ in both
stages. Stage~2 plans typically described placing the top cube back onto an
assumed intact tower, ignoring that all blocks may have scattered and need to
be located and restacked from the base upward.

\textbf{Performed better when Stage~1 constraint was simple geometry.}
Gap-funnel and sheltered-grasp were the clearest successes. For gap-funnel,
Stage~2 scores were high across all models (GPT-5.4-Mini $90\%$, Claude-Haiku
and Claude-Sonnet $100\%$, GPT-5.5 $100\%$), closely matching their Stage~1
scores. The geometric constraint; align sphere with gap, push through, verify, 
translated directly into an ordered sequence of actions. Sheltered-grasp
similarly maintained high Stage~2 scores ($85$--$100\%$), because the physical
shelter forced a specific entry order that was hard to omit. Ramp-sort also
showed good Stage~1 to Stage~2 retention for stronger models, with
Claude-Sonnet and GPT-5.5 both at $90\%$ in Stage~2, consistent with their
Stage~1 scores. 

\textbf{Sequencing is the core requirement of Stage~2.}
The majority of Stage~2 errors stemmed from incorrect or incomplete step
ordering rather than from missing object references. In tool-retrieve, releasing
the hook must come before grasping the cube. The models predicted to use the tool iteself for picking and then releasing the tool.  In stack-collapse, locating
scattered blocks must precede reconstruction. In slide-catch, tilt
tracking and release prediction must precede the catch. If the sequence is wrong,
the plan fails even when all objects and constraints are named. Across the board,
Stage~2 succeeds when each Stage~1 constraint is converted into a clearly
ordered action step, not merely restated in the plan.

\subsection{Stage 3 Analysis: Performance of IMBAgent}
\label{app:insights_stage3}

Tasks requiring precise geometric alignment showed consistent reasoning failures.
In \task{recover-peg-insert}, the agent failed to identify the correct insertion
axis and often rotated in the wrong direction even with privileged object-centric
information (GPT-5.5 $0\%$, GPT-5.5-obj $0\%$). The agent confused the hole
orientation and never achieved alignment. Similarly, in \task{balance-medium},
the agent partially understood the task by grasping near the rod's center, but
reduced execution to a simple pick-and-place behavior instead of true balancing
(GPT-5.5 $0\%$, GPT-5.5-obj $0\%$). Failed placements were never corrected or
retried, showing limited online recovery.

Tasks that only required approximate interactions showed strong performance.
In \task{domino-single}, the agent repeatedly attempted corrective pushes after
failures and eventually succeeded in collapsing the domino chain in all runs
(GPT-5.5 $100\%$, GPT-5.5-obj $100\%$). A similar trend appeared in
\task{domino-select}, where privileged object-centric information removed
ambiguity about the target chain (GPT-5.5 $50\%$, GPT-5.5-obj $100\%$). With
accurate pose information, the agent reliably aligned with the correct domino
sequence and achieved high success rates.

Across several tasks, policies followed open-loop plans and rarely adapted after
execution errors. In \task{pick-place-ycb}, the agent continued executing the
initial plan even after failed grasps or incorrect placements, often treating
the task as complete regardless of the final outcome (GPT-5.5 $0\%$,
GPT-5.5-obj $0\%$). These failures were amplified by irregular YCB object
geometries, where unstable grasps required corrective adjustments that never
emerged.

\textbf{Privileged information can both help performance and hurt reasoning.}
Object-centric information improved performance in tasks where ambiguity was
mainly spatial, but hurt reasoning about transformations or hidden structure.
In \task{domino-select}, exact pose information enabled reliable target
identification and directional pushing (GPT-5.5 $50\%$, GPT-5.5-obj $100\%$).
In contrast, \task{mirror-pick-place} exposed failures in reasoning about
mirrored coordinate frames (GPT-5.5 $30\%$, GPT-5.5-obj $100\%$). In
image-only settings, the agent sometimes succeeded by visually grounding the
mirrored scene and correcting actions online. However, with privileged poses,
the agent ignored image observations and directly executed actions toward the
reported coordinates without transforming them into the mirrored frame.

\textbf{Policies fail to infer latent interaction strategies.}
Several failures came not from low-level control, but from failing to infer
the correct physical strategy. In \task{cup-extract}, the agent never
discovered that rotating the cup was necessary to release the cube inside
(GPT-5.5 $0\%$, GPT-5.5-obj $0\%$). Instead, it often grasped the rim rather
than the handle, blocking the object from exiting. Similarly, in
\task{occluder-push}, the agent often failed to infer that the cube could be
hidden beneath the plates and prematurely terminated execution, treating the
task as impossible (GPT-5.5 $0\%$, GPT-5.5-obj $0\%$). Even when the plates
were identified as occluders, the generated actions failed to remove them
effectively. With privileged object poses, the agent bypassed occluder
reasoning entirely and attempted direct interaction with the hidden target,
again resulting in failure.

\subsection{End-to-end Visuomotor Policies}
\label{app:insights_vismotor}

We discuss task-specific behaviors, emergent properties of the policies, their
success and failure modes, implications of these behaviors, and the physical
reasoning axes that current policies fail to capture. All values in the section below are referred from the table \ref{app:per-task-policy} for the Finetuned values. 

\paragraph{Coarse-contact tasks (\textit{domino-select}, \textit{domino-single},
\textit{seesaw-balance}).}
Tasks that require only coarse directional interaction, rather than precise
end-effector placement, achieve the highest success rates across all evaluated
policies (\textit{domino-single}: $\pi_{0.5}$ $73\%$, GR00T $25\%$, DP $70\%$;
\textit{domino-select}: $\pi_{0.5}$ $32\%$, GR00T $17\%$, DP $80\%$). We
observed that even approximately correct behaviors can solve the domino tasks,
sometimes by directly knocking intermediate dominoes instead of propagating
through the full chain. A similar pattern appears in \textit{domino-select},
where policies learn an approximate spatial correspondence for selecting the
correct chain. The motion remains coarse, but the correspondence is learned
reliably. On \textit{seesaw-balance}, policies failed to exhibit emergent
corrective behavior ($\pi_{0.5}$ $10\%$, GR00T $0\%$, DP $13\%$). Unlike
humans in teleoperation demonstrations, the policies did not repeatedly adjust
load placement until the beam reached equilibrium.

For tasks requiring accurate end-effector placement within a narrow success
region, policies often demonstrate qualitatively correct motion plans but fail
at the geometry level. On \textbf{\textit{balance-hard}}, the object is
typically grasped successfully and placed near the target location ($\pi_{0.5}$
$58\%$, GR00T $3\%$, DP $20\%$). Importantly, the model learns to grasp near
the center of mass, which is crucial for stable manipulation. The policy also
shows adjustment behaviors to better align the center of mass during placement.
However, the harder balance variants require substantially more precise control,
leading to lower success rates (balance-medium: $\pi_{0.5}$ $71\%$, GR00T
$2\%$, DP $28\%$). In \textbf{\textit{ball-cage-escape}}, despite moderate task
success ($\pi_{0.5}$ $45\%$, GR00T $22\%$, DP $43\%$), the policy behavior
differs significantly from human strategies. Human teleoperators first align the
cage into orientations from which rotating the object naturally releases the
ball. Policies fail to consistently learn this strategy, although they still
perform multiple reorientation attempts, some of which succeed. On
\textbf{\textit{keyboard-typing}}, the high-level intent is preserved, but
cumulative end-effector drift causes frequent key misalignments ($\pi_{0.5}$
$13\%$, GR00T $4\%$, DP $32\%$). The correct finger posture often reaches an
adjacent key instead of the target. Furthermore, in nearly $60\%$ of failures,
the policy misunderstands the task semantics and presses keys that were already
pressed.

\paragraph{Dynamic interception and predictive tasks (\textit{bounce-intercept},
\textit{linear-intercept}, \textit{pendulum-grasp}).}
Policies generate trajectories that approximately follow the motion direction of
the target object but fail to accurately match its speed (bounce-intercept:
$\pi_{0.5}$ $0\%$, GR00T $4\%$, DP $0\%$; linear-intercept: $\pi_{0.5}$ $4\%$,
GR00T $4\%$, DP $1\%$). An interesting observation is that successful episodes
primarily occur when the object moves toward the robot arm, while failures are
common when the object moves away. In \textit{pendulum-grasp}, the policy learns
to move toward a small set of likely grasp locations due to the limited number
of feasible interception points ($\pi_{0.5}$ $28\%$, GR00T $8\%$, DP $35\%$).
Common failure modes include grasping too early or too late, indicating partial
understanding of timing, proximity, and reactivity.

The \textbf{\textit{edge-slide}} task fails for most VLAs with a similar failure
mode: the policies apply excessive force while exposing the edge, causing the
object to fall ($\pi_{0.5}$ $7\%$, GR00T $0\%$). Diffusion Policy learns
smoother motion speeds compared to VLAs, but the dominant failure mode remains
similar (DP $28\%$).

\paragraph{Mirror correspondence and adaptation (\textit{mirror-pick-place}).}
\textit{Mirror-pick-place} results in near-zero performance for most VLAs
($\pi_{0.5}$ $0\%$, GR00T $0\%$). This strongly supports our central hypothesis
that humans possess implicit intuitive correspondence reasoning. Human
teleoperators were shown only mirrored observations and, after roughly
$20$--$30$ trials, adapted reliably to the reversed control mapping. Humans
quickly infer correspondences such as ``left becomes right'' and ``up becomes
down.''

This adaptation does not emerge in VLAs, likely because mirrored interaction is
outside the distribution of their pretraining data. In contrast, Diffusion
Policy performs substantially better, achieving nearly $98\%$ success (DP
$98\%$), likely because it learns representations directly from task-specific
demonstrations. Interestingly, our Stage~3 IMBAgent achieves approximately
$30\%$ success and occasionally discovers through iterative interaction that
observation-action correspondences are reversed (GPT-5.5 $30\%$, GPT-5.5-obj
$100\%$). These results highlight that our benchmark evaluates not only physical
understanding but also broader forms of intuitive reasoning.

\paragraph{Compositional organization and packing (\textit{packing},
\textit{ramp-sort}).}
In \textit{packing}, none of the evaluated policies achieve successful
completion ($\pi_{0.5}$ $0\%$, GR00T $0\%$, DP $0\%$). Policies attempt
pick-and-place behaviors but fail to organize subtasks in a way that satisfies
the geometric packing constraints. The objects fit together in only a small set
of valid configurations, enforced by the L-shaped geometry. In \textit{ramp-sort},
we observe two dominant failure modes ($\pi_{0.5}$ $0\%$, GR00T $0\%$, DP
$7\%$). First, policies often fail to place the ball on the correct side of the
ramp. Second, even when placement occurs on the correct side, inaccurate
positioning causes the ball to roll into the wrong bin. As a result, the task
is highly sensitive to action precision.

\paragraph{Bimanual coordination (\textit{hetbi-plate-shelf}, \textit{hetbi-bowl-table}).}
The heterogeneous bimanual tasks are among the most challenging tasks in the
benchmark, as they require simultaneous success across multiple coordination
axes (hetbi-plate-shelf: $\pi_{0.5}$ $0\%$, GR00T $0\%$, DP $0\%$). Failures
are distributed across several stages of execution. Grasping itself is usually
reliable, suggesting that finetuning successfully teaches the policies to acquire
the object. Approximately $80\%$ of failures occur during the handover stage.
Interestingly, a similar trend is observed in the bowl-table task ($\pi_{0.5}$
$27\%$, GR00T $0\%$, DP $0\%$), although some failed handovers accidentally
invert the bowl and still lead to successful upright placement on the table. In
such cases, the task semantics are not followed strictly, but the final goal
state is nevertheless achieved.

\paragraph{Physical stability reasoning (\textit{shape-stack}).}
In \textit{shape-stack}, we observe that the notion of physical stability is not
naturally encoded in current policies ($\pi_{0.5}$ $0\%$, GR00T $0\%$, DP
$0\%$). Human teleoperators quickly realize that certain objects, such as cones,
cannot produce stable stacks unless placed on top. Policies learn the stacking
action itself but fail to reason about long-term physical stability.

\paragraph{Active discovery under occlusion (\textit{occluder-push}).}
Diffusion Policy significantly outperforms VLAs on \textit{occluder-push}
($\pi_{0.5}$ $15\%$, GR00T $0\%$, DP $50\%$). VLA baselines frequently attempt
to directly pick the occluding plates instead of performing clearing motions for
active discovery. Diffusion Policy, trained from scratch on demonstrations,
learns the intended clearing behavior more effectively. The dominant failure mode
for Diffusion Policy occurs when the clearing motion becomes too aggressive,
causing both the occluding plates and the hidden cubes to fall off the table
simultaneously.

Overall, the rollout analysis reveals a consistent gap between high-level intent
and physically precise execution. Policies often learn coarse task structure,
approximate correspondences, and object interaction strategies, but fail when
tasks require accurate geometry, stability reasoning, predictive timing, or
adaptive corrections. Success is strongest in tasks with tolerant dynamics and
broad success regions, while failures are dominated by incorrect physical
reasoning, and limited adaptation where intuitive actions are required.

\section{Miscellaneous}
\label{app:repro}

\paragraph{Hardware.} Training and evaluations reported in \S\ref{sec:experiments} ran on a mix of NVIDIA H100-80GB GPUs (for VLAs), and on L40S (48~GB) for Diffusion policy.

\paragraph{Vision-Language Models APi versions}
Evaluations used the following VLM backbone versions:
\begin{itemize}
    \item \textbf{GPT-5.5} (\texttt{gpt-5.5-2026-04-23})
    \item \textbf{GPT-5.4-mini} (\texttt{gpt-5.4-mini-2026-03-17})
    \item \textbf{Claude Sonnet 4.6} (\texttt{claude-sonnet-4-6})
    \item \textbf{Claude Haiku 4.5} (\texttt{claude-haiku-4-5-20251001})
    \item \textbf{Gemma 4 31B Instruct} (\texttt{google/gemma-4-31B-it})
\end{itemize}
All models were queried via their respective APIs or hosted inference endpoints. No fine-tuning was applied to any VLM backbone. All the models were used with maximum reasoning.

\paragraph{Data}
\begin{itemize}
\item Sample Data: Hosted on \href{https://huggingface.co/imbench}{Hugging Face Datasets}. Individual task datasets can be explored by entering the dataset ID in the official Hugging Face dataset viewer.
\end{itemize}